\documentclass{ieeeaccess}
\usepackage{cite}
\usepackage{amsmath,amssymb,amsfonts}
\usepackage{algorithmic}
\usepackage{graphicx}
\usepackage{textcomp}
\usepackage{url}

\usepackage{comment}
\usepackage{multirow}
\usepackage{threeparttable}
\usepackage{booktabs}
\usepackage[bottom]{footmisc}

\usepackage{bm}
\makeatletter
\AtBeginDocument{\DeclareMathVersion{bold}
\SetSymbolFont{operators}{bold}{T1}{times}{b}{n}
\SetSymbolFont{NewLetters}{bold}{T1}{times}{b}{it}
\SetMathAlphabet{\mathrm}{bold}{T1}{times}{b}{n}
\SetMathAlphabet{\mathit}{bold}{T1}{times}{b}{it}
\SetMathAlphabet{\mathbf}{bold}{T1}{times}{b}{n}
\SetMathAlphabet{\mathtt}{bold}{OT1}{pcr}{b}{n}
\SetSymbolFont{symbols}{bold}{OMS}{cmsy}{b}{n}
\renewcommand\boldmath{\@nomath\boldmath\mathversion{bold}}}
\makeatother

\def\BibTeX{{\rm B\kern-.05em{\sc i\kern-.025em b}\kern-.08em
    T\kern-.1667em\lower.7ex\hbox{E}\kern-.125emX}}

\begin{document}
\history{}
\doi{10.1109/ACCESS.2024.3509714}

\title{Enhancing Biomedical Knowledge Discovery for Diseases: An Open-Source Framework Applied on Rett Syndrome and Alzheimer’s Disease}

\author{\uppercase{Christos Theodoropoulos}\authorrefmark{1},
\uppercase{Andrei Catalin Coman}\authorrefmark{2, 3}, 
\uppercase{James Henderson}\authorrefmark{2}, and
\uppercase{Marie-Francine Moens}\authorrefmark{1}}

\address[1]{LIIR, Department of Computer Science, KU Leuven, 3000 Leuven, Belgium (e-mail: {christos.theodoropoulos, sien.moens}@kuleuven.be)}
\address[2]{Idiap Research Institute, 1920 Martigny, Switzerland (e-mail: {andrei.coman, james.henderson}@idiap.ch)}
\address[3]{EPFL, Department of Electrical Engineering, 1015 Lausanne, Switzerland}

\tfootnote{ This work is supported in part by the Research Foundation – Flanders (FWO) and the Swiss National Science Foundation (SNSF) under Grants G094020N and 200021E\_189458. Christos Theodoropoulos and Marie-Francine Moens are affiliated with Leuven.AI - KU Leuven Institute for AI, B-3000, Leuven, Belgium.}

\markboth
{Christos Theodoropoulos \headeretal: Enhancing Biomedical Knowledge Discovery for Diseases}
{Christos Theodoropoulos \headeretal: Enhancing Biomedical Knowledge Discovery for Diseases}

\corresp{Corresponding author: Christos Theodoropoulos (christos.theodoropoulos@kuleuven.be)}

\begin{abstract}
The rapidly increasing number of biomedical publications presents a significant challenge for efficient knowledge discovery. To address this, we introduce an open-source, end-to-end framework designed to automatically extract and construct knowledge about specific diseases directly from unstructured text. To facilitate research in disease-related knowledge discovery, we create two annotated datasets focused on Rett syndrome (RS) and Alzheimer's disease (AD), enabling the identification of semantic relations between various biomedical entities. We perform extensive benchmarking to evaluate different approaches for representing relations and entities, providing insights into optimal modeling strategies for semantic relation detection and highlighting language models' competence in knowledge discovery. To gain a deeper understanding of the internal mechanisms of transformer models, we also conduct probing experiments, analyzing different layer representations and attention scores, to explore transformers' ability to capture semantic relations within the text. Both the code and the datasets will be publicly available\footnote{\url{https://github.com/christos42/Enhancing-Biomedical-Knowledge-Discovery-for-Diseases}}, encouraging further research in biomedical knowledge discovery.
\end{abstract}

\begin{keywords}
Benchmarking, corpus creation, evaluation, knowledge discovery, language model probing, language resources, NLP datasets, relation detection. 
\end{keywords}

\titlepgskip=-21pt

\maketitle
\section{Introduction}
\label{sec:introduction}

\begin{figure}[!t]
  \centering
  \includegraphics[width=\columnwidth]{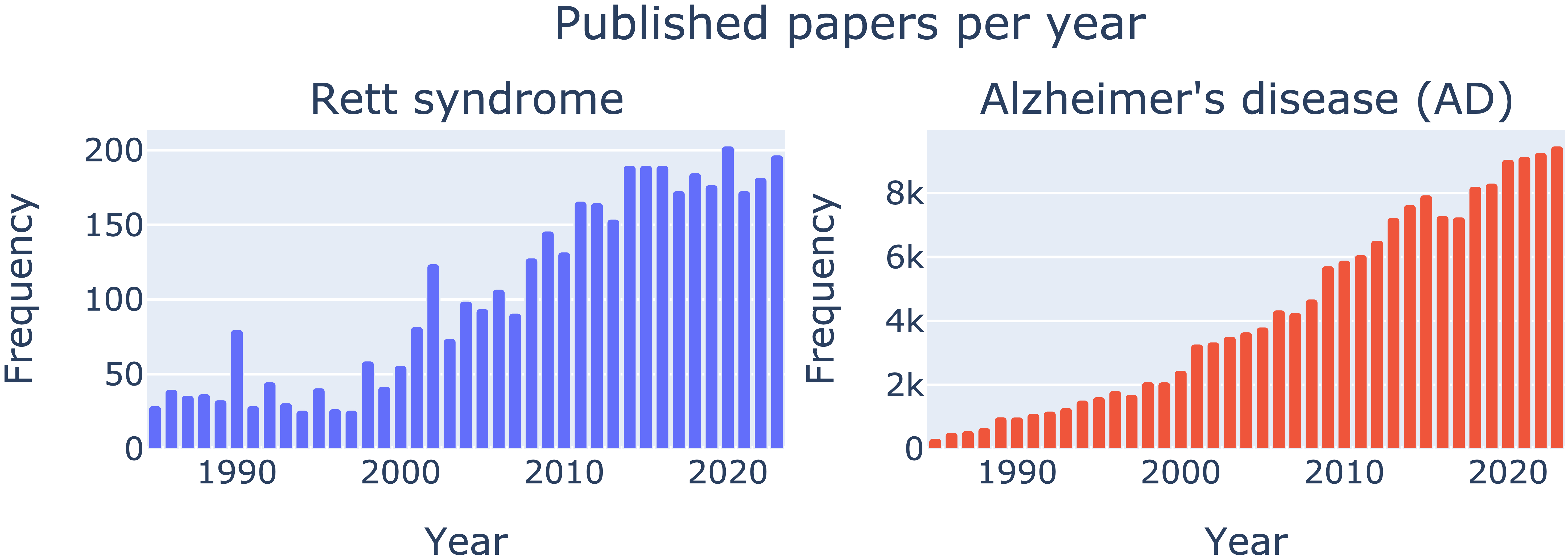}
  \vspace{-7mm}
  \caption{\label{fig:publication_trend}Publication Trends: RS and AD}
  \vspace{-8mm}
\end{figure}

Knowledge discovery \cite{wang2023scientific, shu2023knowledge} is a pivotal research domain due to the surge in publications, which makes keeping up with new findings challenging, necessitating automated knowledge extraction and processing. Of particular concern is biomedical literature, in which updates occur with increasing frequency (Fig. \ref{fig:publication_trend}). Despite advances in healthcare, many diseases, such as Alzheimer's disease (AD) \cite{scheltens2021alzheimer, trejo2023neuropathology}, multiple sclerosis \cite{mcginley2021diagnosis, attfield2022immunology} and muscular dystrophy \cite{duan2021duchenne, bez2023duchenne}, lack effective cures. Additionally, over 1,200 rare disorders have limited or no cures according to the National Organization for Rare Disorders.\footnote{\url{https://rarediseases.org/rare-diseases/}} Discovering new scientific insights from research papers can expedite the understanding of disease and accelerate cure development.

This paper presents an end-to-end framework for detecting medical entities in unstructured text and annotating semantic relations, thereby enabling automated knowledge discovery for diseases. We employ a multi-stage methodology for the data acquisition, annotation, and model evaluation. The process starts with gathering relevant PubMed abstracts from PubMed to form the corpus. Entities are identified and extracted, followed by the co-occurrence graph generation that models the intra-sentence co-occurrence of the entities across the corpus. Leveraging the processed text and co-occurrence graph, an algorithm samples sentences to create gold-standard datasets. Medical experts label the semantic relations between entities within these sentences via an annotation portal. The framework's versatility allows application across various diseases and enables expansion to encompass knowledge about symptoms, genes, and more. This study focuses on two diseases of particular research interest: Rett syndrome (RS) \cite{petriti2023global}  and AD. These diseases are selected due to their significant impact and the absence of a cure, highlighting the urgency for advancements in understanding and treatment. We introduce two curated datasets tailored to detect semantic relations between entities in biomedical text related to RS and AD. The datasets are used for benchmarking, testing techniques for representing relations and entities, and assessing language models' capabilities in knowledge discovery. This work probes the layer outputs of transformer models  \cite{vaswani2017attention} and their attention patterns to reveal their ability to implicitly capture semantic relations in biomedical text.

RS \cite{sandweiss2020advances} poses challenges due to its sporadic nature and rare expression across diverse racial groups. The elusive nature of this disorder undermines its comprehension and stresses the pressing need for a cure. Rare diseases collectively affect a substantial portion of the population, with over 30 million people affected in Europe alone \cite{PAKTER2024100018}. AD is characterized by its prevalence among older populations, with millions of patients worldwide as it is the most common type of dementia (60-70\% of cases) \cite{alzheimers2024report}. With life expectancy on the rise, the projected increase in Alzheimer’s disease's cases accentuates the urgency of finding a cure.

In summary, the key paper's contributions are:
\begin{itemize}
    \item Development of an open-source end-to-end framework to build disease knowledge directly from raw text.
    \item Two annotated datasets for RS and AD provide gold labels for semantic relations, aiding disease knowledge discovery research.
    \item 
    Benchmarking on the datasets examines methods for relation and entity representation, offering insights into optimal approaches for semantic relation detection and emphasizing the knowledge discovery capabilities of language models.
    \item Probing experiments with different layer representations and attention scores assess transformers' inherent ability to capture semantic relations.
\end{itemize}

\section{Related Work}
\label{sec:related_work}

\noindent\textbf{Information Extraction Datasets.} Several biomedical datasets aim to enhance Information Extraction (IE) system development \cite{theodoropoulos-etal-2021-imposing, nasar2021named, detroja2023survey, huang2024surveying}, typically focusing on one or a few entity types and their interactions. AIMed \cite{bunescu2005comparative}, BioInfer \cite{pyysalo2007bioinfer}, and BioCreative II PPI IPS \cite{krallinger2008overview} formulate protein-protein interactions. The chemical-protein and chemical-disease interactions are modeled by DrugProt \cite{miranda2021overview} and BC5CDR \cite{li2016biocreative}, respectively. ADE \cite{gurulingappa2012development}, DDI13 \cite{herrero2013ddi}, and n2c2 2018 ADE \cite{henry20202018} include drug-ADE (adverse drug effect) and drug-drug interactions. EMU \cite{doughty2011toward}, GAD \cite{bravo2015extraction} and RENET2 \cite{su2021renet2} contain relations between genes and diseases. N-ary \cite{peng-etal-2017-cross} incorporates drug-gene mutation interactions. The event extraction task is illustrated by GE09 \cite{kim-etal-2009-overview}, GE11 \cite{kim-etal-2011-overview-genia}, and CG \cite{pyysalo-etal-2013-overview}. DDAE \cite{lai2019using} includes disease-disease associations. BioRED \cite{luo2022biored} focuses on document-level relations for various entities. Unlike these datasets, the datasets of our paper focus on RS and AD, including entities of up to 82 different semantic types, and model the semantic relation between them.

\noindent\textbf{Knowledge Discovery.} \cite{gottlieb2011predict} presents PREDICT, a method for ranking potential drug-disease associations to predict drug indications. \cite{romano2024alzheimer} releases AlzKB, a heterogeneous graph knowledge base for AD, constructed using external data sources and describing various medical entities (e.g., chemicals and genes). Other graph-based efforts model knowledge around AD for tasks such as drug repurposing \cite{daluwatumulle2022silico, nian2022mining, hsieh2023synthesize}, gene identification \cite{binder2022machine}, or as general knowledge repositories \cite{sugis2019hena}. Recently, the study by \cite{barriuso2024recommendation} focuses on a recommendation system that suggests medical articles based on diagnoses in discharge summaries, which involves semantic similarity modeling between summaries and scientific publications. \cite{feng2024semantic} introduces MLTE, a feature fusion approach for semantic textual similarity analysis, which aims to measure the similarity between pairs of clinical texts. 

Another paradigm for knowledge discovery is the open information extraction (OIE) setup \cite{etzioni2008open, mausam-etal-2012-open}, which faces challenges such as data consistency, performance evaluation, and semantic drift \cite{ijcai2022p793}. Research efforts \cite{nebot2011semantics, movshovitz2012bootstrapping, nebot2014exploiting, de2017discovering, wang2018open} aim to address these problems and extract knowledge with little or no supervision. Advances in literature-based discovery \cite{gopalakrishnan2019survey, thilakaratne2019systematic} try to identify novel medical entity relations using graph-based \cite{kilicoglu2020broad, nicholson2020constructing}, machine learning \cite{zhao2021recent, lardos2022computational}, and co-occurrence methods \cite{kuusisto2020kinderminer, millikin2023serial}. \cite{tian2024opportunities} stresses the potential of large language models (LLMs) to summarize, simplify, and synthesize medical evidence \cite{peng2023ai, tang2023evaluating, shaib-etal-2023-summarizing}, suggesting that LLMs may have encoded biomedical knowledge \cite{singhal2023large}. To exploit this potential, we explore the construction of LM representations for knowledge discovery. To the best of our knowledge, no systematic approach assembles knowledge about RS. Unlike previous work, we introduce a, in principle, disease-agnostic framework, to acquire knowledge about RS and AD starting from raw text.

\section{Data Pipeline}
\label{sec:data_pipeline}

\begin{figure}[!t]
  \centering
  \includegraphics[width=\columnwidth]{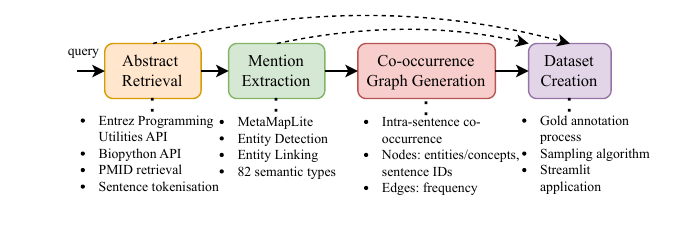}
  \caption{\label{fig:data_pipeline}The pipeline starts with abstract retrieval using a natural language query. Next, entities are detected and linked to UMLS, followed by the co-occurrence graph generation. The final step is the dataset creation using the processed text (abstract retrieval and mention extraction steps) and co-occurrence graph.}
  \vspace{-2mm}
\end{figure}

We focus on developing a robust data pipeline (Fig. \ref{fig:data_pipeline}) to annotate sentences with entities associated with the Unified Medical Language System (UMLS) \cite{bodenreider2004unified, elkin2023unified}. 
The first step involves the retrieval of textual abstracts, followed by mention extraction, which includes entity detection and linking to UMLS. We construct a co-occurrence graph to highlight the interconnections between the entities in the text. The processed text and co-occurrence graph are then used to develop two curated datasets with precise entity annotations and semantic relations between the detected entity pairs. Appendix \ref{sec:data_pipeline_appendix} provides additional information on the data pipeline. We state that the selection of the open-source tools is based on their wide adoption by the research community, their accessibility, and their suitability for biomedical text processing.

\noindent
\textbf{Abstract retrieval.} We retrieve the PubMed\footnote{\url{https://pubmed.ncbi.nlm.nih.gov/}} articles ids based on a query (e.g., Rett syndrome) and extract their open-access abstracts. To accomplish this, we leverage the official Entrez Programming Utilities \cite{kans2024entrez} and the Biopython API \cite{cock2009biopython} (BSD 3-Clause License), ensuring access to a vast repository of biomedical literature. After obtaining the PubMed IDs (PMIDs), we retrieve abstracts from the specified articles and tokenize the text into sentences using NLTK \cite{bird2009natural} (Apache License 2.0). To facilitate effective abstract retrieval, we implement an iterative approach to circumvent the API's limitation of retrieving only 10,000 article IDs per query. This iterative process enables us to access a comprehensive set of PubMed IDs (PMIDs) related to the query. 

\noindent
\textbf{Mention extraction.} MetaMapLite \cite{aronson2001effective, demner2017metamap} (open-source BSD License) is provided by the National Library of Medicine (NLM) for extracting biomedical entities and mapping them to Concept Unique Identifiers (CUIs) within UMLS. The tool is updated every two years to incorporate the latest medical terminology and to ensure its accuracy in extraction and mapping. MetaMapLite simultaneously extracts mentions and links them to UMLS in one step, efficiently associating the mentions with their corresponding CUIs. We detect a diverse range of entities, spanning 82 unique semantic types and covering a broad spectrum of biomedical concepts, including diseases, biologically active substances, anatomical structures, genes, and more (Table \ref{tab:semantic_types}). Detailed entity detection often leads to overlapping or successive entities in text. To address this issue, our pipeline incorporates a heuristic merging strategy that consolidates overlapping or subsequent entities into cohesive units. For example, in the sentence: "To test norepinephrine augmentation as a potential disease-modifying therapy, we performed a biomarker-driven phase II trial of atomoxetine, a clinically-approved \textit{norepinephrine transporter} \textit{inhibitor}, in subjects with mild cognitive impairment due to AD.", the subsequent relevant mentions \textit{norepinephrine transporter} and \textit{inhibitor} are merged into one entity.

\begin{table*}[!ht]
  \caption{\label{tab:semantic_types}List of the 82 semantic types of the MetaMapLite-based pipeline.}
  \resizebox{\textwidth}{!}{
        \begin{tabular}{ccccc}
            \hline
            \multicolumn{5}{c}{\textbf{Semantic Types}}\\
            \hline
            Amino Acid, Peptide, or Protein & Acquired Abnormality & Amino Acid Sequence & Amphibian  & Anatomical Abnormality\\
            \hline
            Animal & Anatomical Structure  & Antibiotic & Archaeon & Biologically Active Substance\\
            \hline
            Bacterium & Body Substance & Body System  & Behavior & Biologic Function \\
            \hline
            Body Location or Region & Biomedical or Dental Material & Body Part, Organ, or Organ Component & Body Space or Junction & Cell Component\\
            \hline
            Cell Function & Cell & Congenital Abnormality  & Chemical & Chemical Viewed Functionally \\
            \hline
            Chemical Viewed Structurally & Clinical Attribute & Clinical Drug & Cell or Molecular Dysfunction & Carbohydrate Sequence \\
            \hline
            Diagnostic Procedure & Daily or Recreational Activity & Disease or Syndrome & Environmental Effect of Humans & Element, Ion, or Isotope \\
            \hline
            Experimental Model of Disease & Embryonic Structure & Enzyme & Eukaryote & Fully Formed Anatomical Structure \\
            \hline
            Fungus & Food & Genetic Function & Gene or Genome & Human-caused Phenomenon or Process \\
            \hline
            Health Care Activity & Hazardous or Poisonous Substance & Hormone & Immunologic Factor & Individual Behavior \\
            \hline
            Inorganic Chemical & Injury or Poisoning & Indicator, Reagent, or Diagnostic Aid & Laboratory Procedure & Laboratory or Test Result \\
            \hline
            Mammal & Molecular Biology Research Technique & Mental Process & Mental or Behavioral Dysfunction & Molecular Sequence \\
            \hline
            Neoplastic Process & Nucleic Acid, Nucleoside, or Nucleotide & Nucleotide Sequence & Organic Chemical & Organism Attribute \\
            \hline
            Organism Function & Organism & Organ or Tissue Function & Pathologic Function & Pharmacologic Substance \\
            \hline
            Plant & Organism & Population Group & Receptor & Reptile \\
            \hline
            Substance & Social Behavior & Sign or Symptom & Tissue & Therapeutic or Preventive Procedure \\
            \hline
            Virus & Vitamin & Vertebrate & & \\
            \hline
        \end{tabular}}
  \vspace{-3mm}
\end{table*}

\begin{figure}[!t]
  \centering
  \includegraphics[width=\columnwidth]{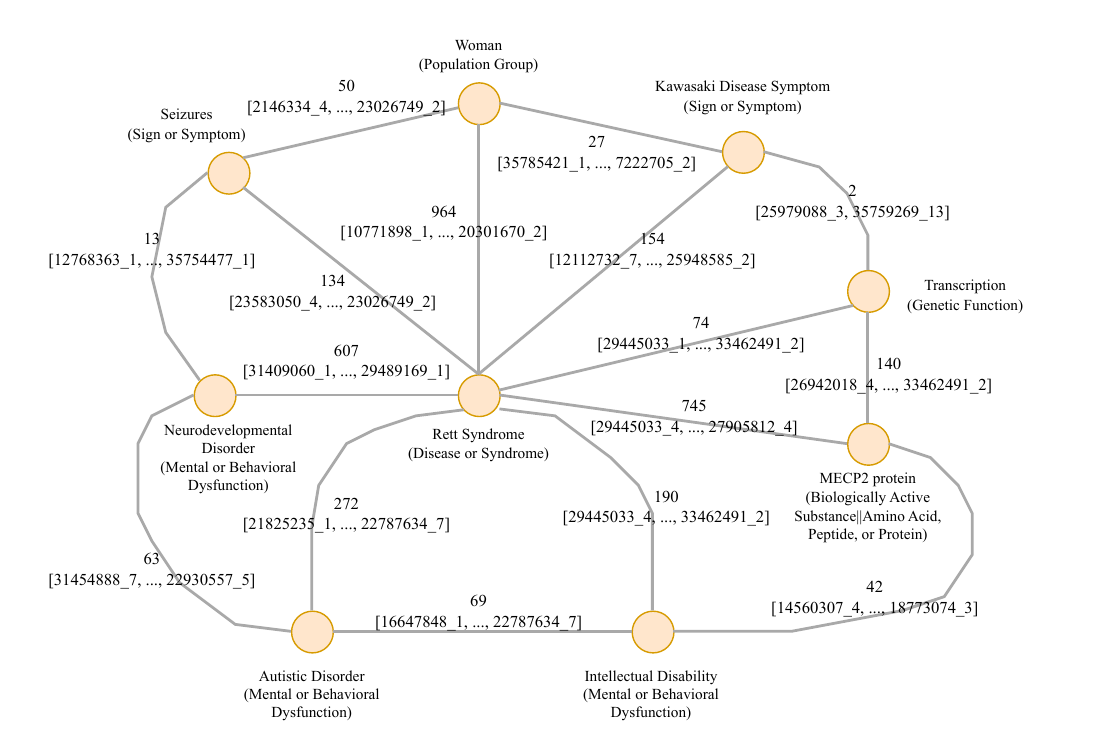}
  \caption{\label{fig:co-occurrence_subgraph}Visualization of a subgraph of the co-occurrence graph for the Rett Syndrome corpus. Each node corresponds to a unique CUI with the related textual description and contains the semantic type. The edge label includes the number of times two entities co-occur in a sentence and a list with the sentence IDs where the connected entities are detected.}
  \vspace{-3mm}
\end{figure}

\noindent
\textbf{Co-occurrence graph generation.} We model the intra-sentence co-occurrence between the entities (an example of a co-occurrence subgraph is presented in Fig. \ref{fig:co-occurrence_subgraph}). Each node in the graph corresponds to a unique CUI and contains metadata including the semantic type and list of sentence IDs where the corresponding entity is detected. An edge between two nodes signifies that corresponding entities co-occur within the same sentence. The edge weight represents the number of times two entities co-occur in a sentence throughout the text corpus.

\noindent
\textbf{Dataset creation.} Leveraging the extracted co-occurrence graph, we define two distinct probability distributions to select sentences for manual annotation. The first distribution $\mathcal{P}$ focuses on common pairs of co-occurred entities, with higher frequency in the co-occurrence graph resulting in a higher likelihood of sampling. The second distribution $\mathcal{IP}$ prioritizes novel/rare pairs of co-occurred entities, selecting sentences where the entities have a lower frequency in the co-occurrence graph. We sample 50\% of sentences using $\mathcal{P}$ and 50\% using $\mathcal{IP}$ to ensure a balance of common and potentially novel pairs of co-occurring entities in the datasets. The sentence sampling algorithm is detailed in Appendix \ref{sec:data_pipeline_appendix}.

Then, we develop an annotation portal using the streamlit\footnote{\url{https://streamlit.io/}} library, providing a user-friendly interface for annotators. Annotators are presented with a sentence containing two highlighted entities and are prompted to categorize the semantic relation between them. Options include \textit{positive} (direct semantic connection), \textit{negative} (negative semantic connection where negative words like "no" and "absence" are present), \textit{complex}, and \textit{no relation} (absence of semantic relation). In collaboration with clinical experts, we identify instances where entities exhibit a semantic connection without a clear positive or negative semantic relation. To capture these complexities, we introduce the "complex" relation type, enhancing the potential for in-depth analysis. The annotation portal offers additional functions such as sentence removal (for non-informative sentences), entity removal (for incorrect entity types or spans), and context addition (for providing additional text to aid in relation type determination). We enlist the expertise of three medical experts to ensure the accuracy and reliability of the annotation process.

\begin{table}[!t]
    \caption{\label{tab:datasets_statistics}Datasets: Statistics of ReDReS and ReDAD and their 
    label distribution.}
    \centering
    \resizebox{\columnwidth}{!}{
        \begin{tabular}{ccccc}
        \hline
        \textbf{Dataset} & \textbf{Sentences} & \textbf{Instances} & \textbf{Unique CUIs} & \textbf{Semantic Types}\\
        \hline
        ReDReS & 601 & 5,259 & 1,148 & 73 \\
        \cmidrule{2-5}
        Train set & 409 (68.1\%) & 3,573 (67.9\%) & 887 & 73 \\
        Dev. set & 72 (12.0\%) & 749 (14.2\%) & 249 & 56 \\
        Test set & 120 (19.1\%) & 937 (17.9\%) & 349 & 57 \\
        \hline
        ReDAD & 641 & 8,565 & 1,480 & 82 \\
        \cmidrule{2-5}
        Train set & 437 (68.2\%) & 5,502 (64.2\%) & 1,114 & 78 \\
        Dev. set & 76 (11.9\%) & 1,188 (13.9\%) & 321 & 60 \\
        Test set & 128 (19.9\%) & 1,875 (21.9\%) & 452 & 58 \\
        \hline
        \multirow{2}{*}{\textbf{Dataset}} & \multicolumn{4}{c}{\textbf{Labels - Type of Relation}} \\
        & \textbf{Positive} & \textbf{Complex} & \textbf{Negative} & \textbf{No Relation} \\
        \hline
        ReDReS & 1,732 (32.9\%) & 1,491 (28.4\%) & 97 (1.8\%) & 1,945 (36.9\%)\\
        \cmidrule{2-5}
        Train set & 1,176 (32.9\%) & 996 (27.9\%) & 69 (1.9\%) & 1,332 (37.3\%)\\
        Dev. set & 241 (32.2\%) & 213 (28.4\%) & 7 (0.9\%) & 288 (38.5\%)\\
        Test set & 313 (33.3\%) & 282 (30.1\%) & 21 (2.2\%) & 321 (34.4\%) \\
        \hline
        ReDAD & 2,496 (29.1\%) & 2,874 (33.6\%) & 125 (1.5\%) & 3,070 (35.8\%)\\
        \cmidrule{2-5}
        Train set & 1,718 (31.2\%) & 1,923 (34.9\%) & 68 (1.2\%) & 1,793 (32.7\%)\\
        Dev. set & 286 (24.1\%) & 373 (32.4\%) & 18 (1.5\%) & 511 (42.0\%)\\
        Test set & 492 (26.2\%) & 578 (30.8\%) & 39 (2.1\%) & 766 (40.9\%)\\
        \hline
    \end{tabular}}
\end{table}

\begin{figure}[!t]
  \centering
  \includegraphics[width=\columnwidth]{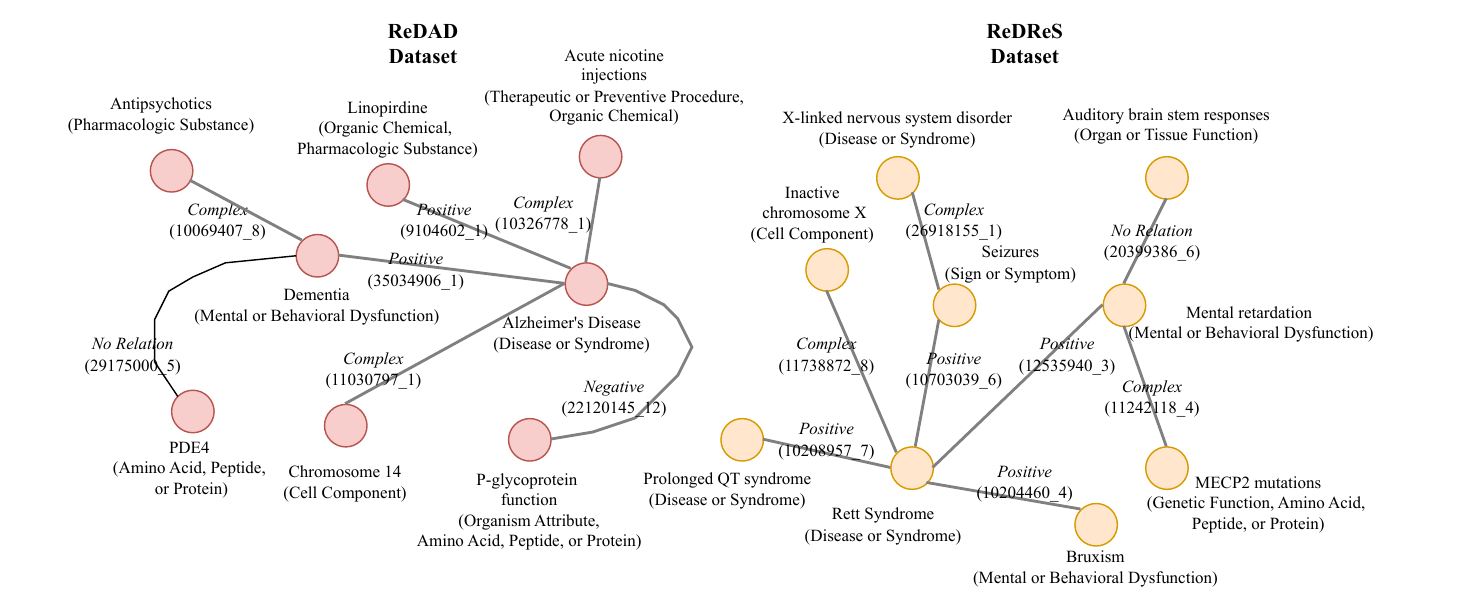}
  \caption{\label{fig:graph_datasets} Graphical example of the ReDAD and ReDReS datasets. Each node corresponds to an entity with a textual description and semantic type. The edge label includes the annotated relation type and the sentence ID where the connected entities are detected.}
\end{figure}

The result of the expert annotation yields two curated datasets. The \textbf{Re}lation \textbf{D}etection dataset for \textbf{Re}tt \textbf{S}yndrome (ReDReS) contains 601 sentences with 5,259 instances and 1,148 unique CUIs (Table \ref{tab:datasets_statistics}). The inter-annotator agreement is measured using the Fleiss kappa score \cite{mchugh2012interrater}, resulting in 0.6143 in the multi-class setup (4 classes) and indicating substantial agreement among annotators \cite{landis1977measurement}. In the binary setup (\textit{relation} or \textit{no relation}), where the classes \textit{positive}, \textit{complex}, \textit{negative} are grouped under the \textit{relation} class, the Fleiss kappa score is 0.7139. The \textbf{Re}lation \textbf{D}etection dataset for \textbf{A}lzheimer's \textbf{D}isease (ReDAD) comprises 641 sentences with 8,565 instances and 1,480 unique CUIs (Table \ref{tab:datasets_statistics}). The Fleiss kappa score is 0.6403 in the multi-class setup and 0.7064 in the binary setup, showing substantial consensus among annotators. The final labels are determined through majority voting, leveraging the labels provided by each expert. Whereas the label distribution across classes is relatively balanced, the negative class is under-represented with 97 and 125 instances in ReDReS and ReDAD respectively (Table \ref{tab:datasets_statistics}). Each dataset is randomly split into train, development, and test sets. A graphical example of the datasets is presented in Fig. \ref{fig:graph_datasets}.

\section{Models}
\label{sec:models}

\begin{figure*}[!t]
  \centering
  \includegraphics[width=\textwidth]{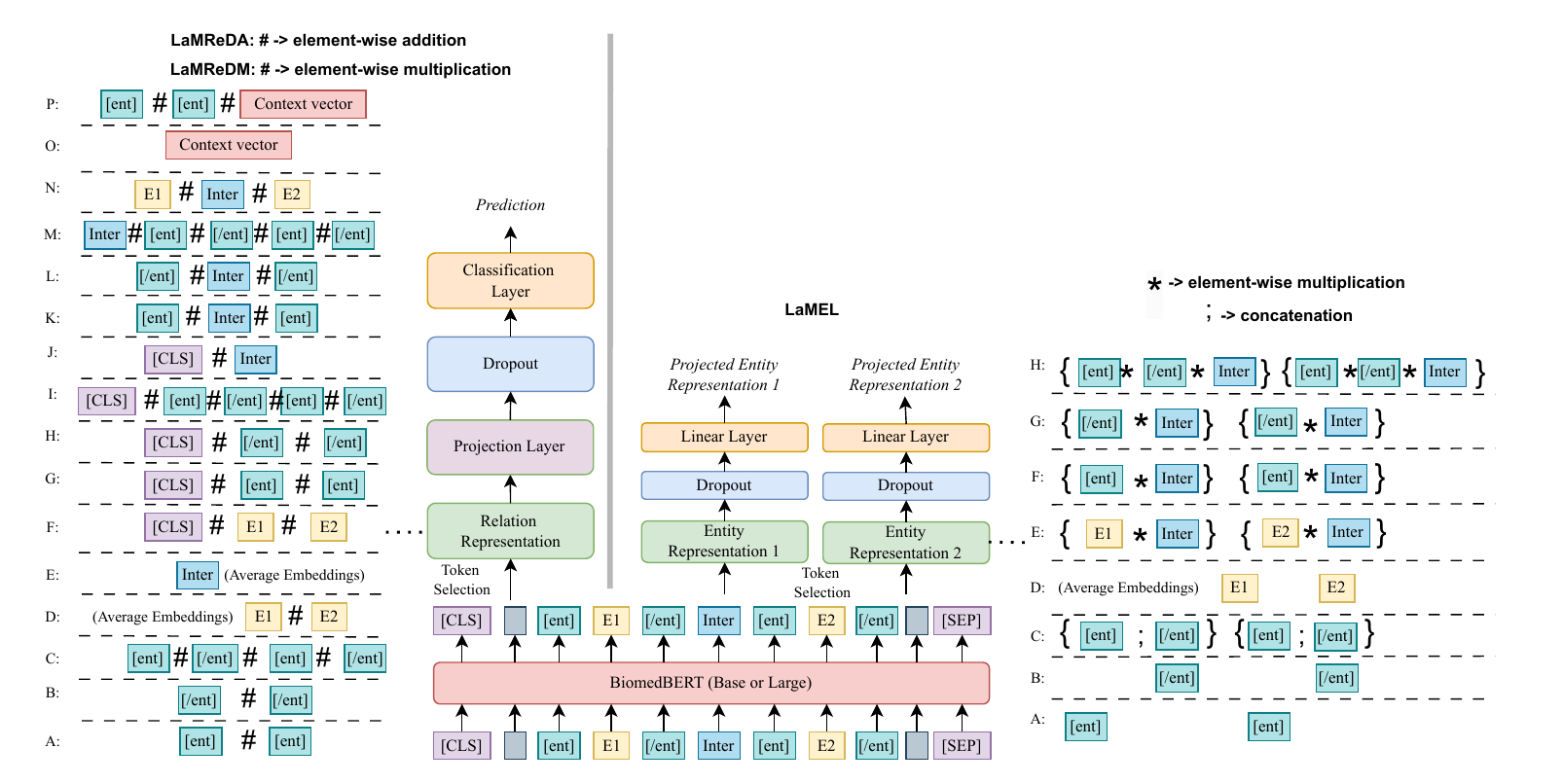}
  \caption{\label{fig:model_architecture}Model Architecture of LaMReDA, LaMReDM (left), and LaMEL (right): Each model encodes the input sequence using BiomedBERT (large or base). For LaMReDA and LaMReDM, different tokens define the relation representation (A-P), passed through a linear projection layer, a dropout layer, and then a classification layer for prediction. The symbol \textit{\#} denotes element-wise addition and multiplication for LaMReDA and LaMReDM, respectively. For LaMEL, different tokens construct the entity representation (A-H), which are sent through a dropout layer and a linear layer to extract the projected entity representations. The symbols \textit{;} and \textit{*} define the concatenation and the element-wise multiplication, respectively, for the LaMEL model.}
\end{figure*}

In this section, we introduce two main models, the \textbf{La}nguage-\textbf{M}odel \textbf{E}mbedding \textbf{L}earning (LaMEL) model and the \textbf{La}nguage-\textbf{M}odel \textbf{Re}lation \textbf{D}etection (LaMReD) model (Fig. \ref{fig:model_architecture}), to benchmark datasets and establish robust baselines. 

\noindent\textbf{Task formulation.} Given a sentence containing two identified entities $e1$ and $e2$, we predict the semantic relation $sem_r$ between them. In the multi-class setup, the labels are: \textit{positive}, \textit{negative}, \textit{complex}, and \textit{no relation}. In the binary setup, the goal is to determine if any relation exists. Special tokens [ent] and [/ent] mark the start and end of each entity within the sentence, ensuring consistent identification and processing of entity boundaries.

\subsection{LaMEL model}

LaMEL (Fig. \ref{fig:model_architecture}) learns an embedding space optimized for relation detection (Fig. \ref{fig:model_architecture}). As the backbone language model (LM), we opt for BiomedBERT \cite{gu2021domain, tinn2023fine} (MIT License), which is available in both uncased base and uncased large versions, accessed via HuggingFace's Transformers library \cite{wolf-etal-2020-transformers}. BiomedBERT is pre-trained on the PubMed corpus, making it well-suited for our task as the curated datasets consist of sentences of abstracts from PubMed papers. Leveraging BiomedBERT ensures that the model can capture language patterns prevalent in biomedical texts. Following LM encoding, we construct the representation of each entity by extracting its contextualized embedding $E_i$ corresponding to each entity $e_i$ from the encoded sequence. Subsequently, entity representations are projected onto the embedding space using a linear layer without changing the embedding dimension. The final prediction is based on the cosine similarity between the two projected entity representations. If the cosine similarity exceeds a predefined threshold, the model predicts a semantic relation between the two entities. This approach leverages the geometric properties of the embedding space to determine the semantic similarity between entities. 
We experiment with diverse strategies for learning entity representations (Fig. \ref{fig:model_architecture}), aiming to optimize the effectiveness of the embedding space for the relation detection task. The explored types of entity representation $E$ are:

\begin{itemize}
    \item \textbf{A, B, C - Special Tokens}: 
        \vspace{-2mm}
        \begin{equation}
            \label{eq:e_a}
            E_{A} = t_{[ent]},
        \end{equation}
        \begin{equation}
            \label{eq:e_b}
            E_{B} = t_{[/ent]},
        \end{equation}
        \begin{equation}
            \label{eq:e_c}
            E_{C} = t_{[ent]} ; t_{[/ent]},
        \end{equation}
    \item \textbf{D - Entity Pool}:
        \vspace{-2mm}
        \begin{equation}
            \label{eq:e_d}
            E_{D} = [t_{E}],
        \end{equation}
    \item \textbf{E - Entity \& Middle Pool}: 
        \vspace{-2mm}
        \begin{equation}
            \label{eq:e_e}
            E_{E} = [t_{E}] * [t_{Inter}],
        \end{equation}
    \item \textbf{F, G, H - Special Tokens \& Middle Pool}:
        \vspace{-2mm}
        \begin{equation}
            \label{eq:e_f}
            E_{F} = t_{[ent]} * [t_{Inter}],
        \end{equation}
        \begin{equation}
            \label{eq:e_g}
            E_{G} = t_{[/ent]} * [t_{Inter}],
        \end{equation}
        \begin{equation}
            \label{eq:e_h}
            E_{H} = t_{[ent]} * t_{[/ent]} * [t_{Inter}],
        \end{equation}
\end{itemize}

\noindent
where $\{E_{A}, E_{B}, E_{D}, E_{E}, E_{F}, E_{G}, E_{H}\} \in \mathbb{R}^d$ and $E_{C} \in \mathbb{R}^{2d}$, $d$ is the embedding size of BiomedBERT base (768) and BiomedBERT large (1024), $;$ defines the concatenation, $*$ holds for the element-wise multiplication, $t_{[ent]}$, $t_{[/ent]}$ are the embeddings of the start and end special tokens of the entities, $[t_{E}]$ and $[t_{Inter}]$ are the averaged pooled representation of the entities and the intermediate tokens between the entities respectively. For example, in the sentence: "[CLS] [ent] \textit{Amyloid fibrils} [/ent] are found in many fatal neurodegenerative diseases such as Alzheimer’s disease, Parkinson’s disease, [ent] \textit{type II diabetes} [/ent], and prion disease. [SEP]", the two identified entities are \textit{amyloid fibrils} and \textit{type II diabetes}. The averaged pooled representation of the two entities corresponds to $[t_{E}]$ independently and $[t_{Inter}]$ is the averaged pooled representation of the intermediate tokens ("are found in many fatal neurodegenerative diseases such as Alzheimer’s disease, Parkinson’s disease").

Fig. \ref{fig:model_architecture} illustrates the model architectures of the LaMEL model (on the right side). The model encodes the input sequence with the large or base version of BiomedBERT model. Different tokens (marked as "A-H") are selected to construct the entity representations. Next, the entity representations pass through a dropout layer and then a linear layer to extract the projected representations. The ";" symbol indicates the concatenation of tokens, while "*" represents element-wise multiplication between specific tokens.

\subsection{LaMReD model}

LaMReD provides two variations that differ in information synthesis (Fig. \ref{fig:model_architecture}), aiming to explore the potential effects of different aggregations \cite{theodoropoulos2023information}. LaMReDA (Fig. \ref{fig:model_architecture}) utilizes element-wise addition to aggregate the entities' representations, whereas LaMReDM (Fig. \ref{fig:model_architecture}) employs element-wise multiplication. The input text is encoded using BiomedBERT (base or large). Following LM encoding, we construct the relation representation by sampling and aggregating tokens from the input sequence. This step enables the model to capture essential features and contextual information relevant to semantic relation classification. To mitigate the risk of overfitting and enhance model generalization, we incorporate a dropout layer \cite{srivastava2014dropout} with a probability of 0.3. The linear classification layer uses the aggregated representation and outputs the predicted label. 

Following the paradigm proposed by \cite{baldini-soares-etal-2019-matching} and \cite{hogan2021abstractified}, we experiment with various approaches for learning relation representations tailored to the relation detection task to empirically ascertain the effectiveness of each strategy (Fig. \ref{fig:model_architecture}). The explored types of relation representation \textit{R} are as follows: 

\begin{itemize}
    \item \textbf{A, B, C - Special Tokens}:
        \begin{equation}
            \label{eq:r_a}
            R_{A} = f(l(t_{[ent]_1}), l(t_{[ent]_2})),
        \end{equation}
        \begin{equation}
            \label{eq:r_b}
            R_{B} = f(l(t_{[/ent]_1}), l(t_{[/ent]_2})),
        \end{equation}
        \begin{equation}
            \label{eq:r_c}
            \begin{aligned}
                 R_{C} = f(l(t_{[ent]_1}), l(t_{[/ent]_1}), \\ l(t_{[ent]_2}), l(t_{[/ent]_2})),
            \end{aligned}
        \end{equation}
    \item \textbf{D - Entity Pool}:
        \begin{equation}
            \label{eq:r_d}
            R_{D} = f(l([t_{E1}]), l([t_{E2}])),
        \end{equation}  
    \item \textbf{E - Middle Pool}:
        \begin{equation}
            \label{eq:r_e}
            R_{E} = l([t_{Inter}]),
        \end{equation} 
    \item \textbf{F - [CLS] token \& Entity Pool}:
        \begin{equation}
            \label{eq:r_f}
            R_{F} = f(l(t_{[CLS]}), l([t_{E1}]), l([t_{E2}])),
        \end{equation}
    \item \textbf{G, H, I - [CLS] token \& Special Tokens}:
        \begin{equation}
            \label{eq:r_g}
            \hspace{-4mm} R_{G} = f(l(t_{[CLS]}), l(t_{[ent]_1}), l(t_{[ent]_2})),
        \end{equation}
        \begin{equation}
            \label{eq:r_h}
            \hspace{-6mm} R_{H} = f(l(t_{[CLS]}), l(t_{[/ent]_1}), l(t_{[/ent]_2})),
        \end{equation}
        \begin{equation}
            \label{eq:r_i}
            \begin{aligned}
            \hspace{-4mm} R_{I} = f(l(t_{[CLS]}), l(t_{[ent]_1}), l(t_{[/ent]_1}), \\ \vspace{-3mm} l(t_{[ent]_2}), l(t_{[/ent]_2})),
            \end{aligned}
        \end{equation}
    \item \textbf{J - [CLS] token \& Middle Pool}:
        \begin{equation}
            \label{eq:r_j}
            R_{J} = f(l(t_{[CLS]}), l([t_{Inter}])),
        \end{equation}
    \item \textbf{K, L, M - Special tokens \& Middle Pool}:
        \begin{equation}
            \label{eq:r_k}
            \hspace{-6mm} R_{K} = f(l(t_{[ent]_1}), l([t_{Inter}]), l(t_{[ent]_2})),
        \end{equation}
        \begin{equation}
            \label{eq:r_l}
            \hspace{-7mm} R_{L} = f(l(t_{[/ent]_1}), l(t_{Inter}]), l(t_{[/ent]_2})),
        \end{equation}
        \begin{equation}
            \label{eq:r_m}
            \begin{aligned}
            \hspace{-6mm} R_{M} = f(l(t_{[ent]_1}), l(t_{[/ent]_1}), l([t_{Inter}]), \\ \vspace{-3mm} l(t_{[ent]_2}), l(t_{[/ent]_2})),
            \end{aligned}
        \end{equation}
    \item \textbf{N - Entity \& Middle Pool}:
        \begin{equation}
            \label{eq:r_n}
            \hspace{-4mm} R_{N} = f(l([t_{E1}]), l([t_{Inter}]), l([t_{E2}])),
        \end{equation}    
    \item \textbf{O, P - Context Vector \& Entity Pool}:
        \begin{equation}
            \label{eq:r_o}
            R_{O} = l(cv),
        \end{equation}    
        \begin{equation}
            \label{eq:r_p}
            R_{P} = f(l([t_{E1}]), l([t_{E2}]), l(cv)),
        \end{equation}    
\end{itemize}

\noindent
where $\{R_{A}, R_{B}, R_{C}, R_{D}, R_{E}, R_{F}, R_{G}, R_{H}, R_{I}, R_{J}, R_{K}, R_{L}, R_{M}, \\ R_{N}, R_{O}, R_{P}\} \in \mathbb{R}^d$, d is the embedding size of BiomedBERT base (768) and BiomedBERT large (1024), $f()$ is the aggregation function, element-wise addition for LaMReDA and element-wise multiplication for LaMReDM,  $l()$ is a linear projection layer with dimension equal to the embedding size, $t_{[ent]_1}$, $t_{[/ent]_1}$, $t_{[ent]_2}$, and $t_{[/ent]_2}$ are the embeddings of the start and end special tokens of the first and second entity and $t_{[CLS]}$ is the representation of the special token [CLS]. We define the averaged pooled representation of the entities and the intermediate tokens between the entities as $[t_{E1}]$, $[t_{E2}]$, and $[t_{Inter}]$, correspondingly. For example, in the sentence: "[CLS] [ent] \textit{Amyloid fibrils} [/ent] are found in many fatal neurodegenerative diseases such as Alzheimer’s disease, Parkinson’s disease, [ent] \textit{type II diabetes} [/ent], and prion disease. [SEP]", the two identified entities are \textit{amyloid fibrils} ($[t_{E1}]$) and \textit{type II diabetes} ($[t_{E2}]$). The averaged pooled representation of the intermediate tokens ("are found in many fatal neurodegenerative diseases such as Alzheimer’s disease, Parkinson’s disease") corresponds to $[t_{Inter}]$. The embeddings of the start and end special tokens of the first (\textit{Amyloid fibrils}) and second (\textit{type II diabetes}) entities are $t_{[ent]_1}$, $t_{[/ent]_1}$, $t_{[ent]_2}$, and $t_{[/ent]_2}$, respectively. In equations \eqref{eq:r_o} and \eqref{eq:r_p}, we utilize the localized context vector $cv$ (additional information is provided in Appendix \ref{sec:context_vector_appendix}) which utilizes the attention heads to locate relevant context for the entity pair and was introduced in ATLOP \cite{zhou2021document}, a state-of-the-art model in document-level relation extraction. 

Fig. \ref{fig:model_architecture} illustrates the model architectures of the LaMReDA and LaMReDM models (on the left side). The models start by encoding the input sequence with the BiomedBERT model, which can be either the large or base version. Specific tokens in the input sequence are designated to define the relation representation (labeled as "A-P"). These relation representations are passed through a linear projection layer, followed by a dropout layer. Finally, a classification layer performs the relation prediction. The figure uses the symbol "\#" to denote element-wise addition and element-wise multiplication for LaMReDA and LaMReDM, respectively.

\subsection{Experimental setup}
\label{subsec:experimental_setup}
The models are trained for 50 epochs and the best checkpoints are retained based on the performance on the development set, measured using the F1-score. We utilize the Adam \cite{kingma2014adam} optimizer with a learning rate of 10\textsuperscript{-5}. The batch size is set to 16 (Table \ref{tab:experimental_parameters}). We conduct experiments in two distinct setups. In the multi-class setup, we evaluate performance using micro and macro F1-score, considering four relation types: \textit{positive}, \textit{negative}, \textit{complex}, and \textit{no relation}. In the binary setup, the objective is to predict whether a relation exists and we assess performance using the micro F1-score. LaMEL is specifically designed for the binary setup. We utilize the official splits of ReDReS and ReDAD (Table \ref{tab:datasets_statistics}) and repeat the experiments 10 times with different seeds (random numbers that are selected to initialize various processes and pseudorandom number generators, such as the random shuffle of the train set before the beginning of a training epoch or the CUDA initialization, ensuring that the results are reproducible). The following numbers are used: 42, 3, 7, 21, 77, 24, 69, 96, 44, 11. The repetition of the experiments is a common practice in the research community to ensure the robustness of the results. We also employ a 5-fold cross-validation approach to ensure the robustness of results. To explore the cross-disease capabilities of our approach, we train the models using one dataset (e.g., ReDReS) and evaluate on the other (e.g., ReDAD), and vice versa. We utilize the relation representation $R_A$ \eqref{eq:r_a} for LaMReDA and LaMReDM and the entity representation $E_A$ \eqref{eq:e_a} for LaMEL. These experiments are repeated 10 times with different seeds, and 15\% of the training data is excluded to define the development set. We use a single NVIDIA RTX 3090 GPU of 24GB to execute the experiments. 

The cross-entropy loss function is used to train LaMReDA and LaMReDM. For LaMEL, the following cosine embedding loss function is used:

\begin{equation}
    \label{eq:loss_function}
    \resizebox{.85\hsize}{!}{$
    \hspace{-3mm}l(x_{1}, x_{2}, y)= \begin{cases}
                            1 - \mathrm{cos}(x_{1}, x_{2}), & \text{if $y = 1$}\\
                            \mathrm{max}(0, \mathrm{cos}(x_{1}, x_{2}) - m), & \text{if $y = -1$}
                        \end{cases}$},
\end{equation}

\noindent
where $x_{1}$ and $x_{2}$ are the projected representations of the two entities, $y$ is the gold-truth label (1 if the entities are correlated, -1 if they are not), $\mathrm{cos()}$ is the cosine similarity in the embedding space, and $m$ is the margin parameter that is set to 0. In the inference step, the threshold for predicting the presence of relation based on the cosine similarity of the two entity representations is set to 0.5 (Table \ref{tab:experimental_parameters}).

\begin{table}[!ht]
  \centering
  \caption{\label{tab:experimental_parameters}Summary of the experimental parameters.}
        \begin{tabular}{|c|c|}
            \hline
            \textbf{Parameter} & \textbf{Value}\\
            \hline
            Epochs & 50\\
            \hline
            Learning Rate & 10\textsuperscript{-5}\\
            \hline
            Batch Size & 16\\
            \hline
            Margin Parameter $m$ & 0\\
            \hline
            Inference Threshold & 0.5\\
            \hline
        \end{tabular}
\end{table}

\begin{table*}[!ht]
  \caption{\label{tab:LaMReDA_LaMReDM_results}LaMReDA and LaMReDM Results (\%) in binary and multi-class setup (BiomedBERT $\square$: base,$\blacksquare$: large): Each cell (unless cross-disease experiments) shows the average F1-score from 10 runs (original test set) and from 5-fold cross-validation setup.}
  \resizebox{\textwidth}{!}{
  \begin{threeparttable}
        \begin{tabular}{cccc|cc|cc|cc|cc|cc}
            \hline
            \multirow{4}{*}{\textbf{Data}} & \multirow{4}{*}{\textbf{Type}\tnote{1}} & \multicolumn{4}{c}{\textbf{Binary setup}} & \multicolumn{8}{|c}{\textbf{Multi-class setup}}\\
            \cmidrule{7-14}
            & & & & & & \multicolumn{4}{c}{\textbf{Micro Evaluation}} & \multicolumn{4}{|c}{\textbf{Macro Evaluation}} \\
            \cmidrule{3-14}
            & & \multicolumn{2}{c}{\textbf{LaMReDA}} & \multicolumn{2}{|c}{\textbf{LaMReDM}} & \multicolumn{2}{|c}{\textbf{LaMReDA}} & \multicolumn{2}{|c}{\textbf{LaMReDM}} & \multicolumn{2}{|c}{\textbf{LaMReDA}} & \multicolumn{2}{|c}{\textbf{LaMReDM}} \\
            \cmidrule{3-14}
            & & \textbf{F\textsubscript{1}}\textsuperscript{$\square$} & \textbf{F\textsubscript{1}}\textsuperscript{$\blacksquare$} & \textbf{F\textsubscript{1}}\textsuperscript{$\square$} & \textbf{F\textsubscript{1}}\textsuperscript{$\blacksquare$} & \textbf{F\textsubscript{1}}\textsuperscript{$\square$} & \textbf{F\textsubscript{1}}\textsuperscript{$\blacksquare$} & \textbf{F\textsubscript{1}}\textsuperscript{$\square$} & \textbf{F\textsubscript{1}}\textsuperscript{$\blacksquare$} &
            \textbf{F\textsubscript{1}}\textsuperscript{$\square$} & \textbf{F\textsubscript{1}}\textsuperscript{$\blacksquare$} &
            \textbf{F\textsubscript{1}}\textsuperscript{$\square$} & \textbf{F\textsubscript{1}}\textsuperscript{$\blacksquare$} \\
            \hline
            \parbox[t]{2mm}{\multirow{17}{*}{\rotatebox[origin=c]{90}{ReDReS}}} & A & 90.72/89.95 & 90.74/\textbf{90.57} & 90.42/89.15 & \textbf{90.71}/89.53 & 74.49/73.91 & 73.96/75.01 & 74.36/73.31 & 74.35/74.91 & 74.52/74.50 & 73.66/74.48 & 74.30/73.06 & 72.81/75.07 \\
            \cmidrule{2-14}
            & B & 90.40/88.79 & 90.28/89.54 & 90.47/89.33 & 90.06/89.74 & 74.27/74.14 & 73.72/74.79 & 74.26/74.45 & 73.57/75.34 & 74.32/74.15 & 73.65/75.19 & \textbf{74.38}/74.31 & 73.11/75.74 \\
            \cmidrule{2-14}
            & C & 90.85/89.69 & 90.75/89.75 & 90.51/88.84 & 89.14/89.16 & \textbf{74.93}/72.98 & 73.54/74.59 & 74.31/72.71 & 73.69/73.56 & \textbf{74.96}/73.75 & 73.44/74.74 & 74.10/72.83 & 73.49/73.88 \\
            \cmidrule{2-14}
            & D & 90.55/89.29 & \textbf{90.93}/89.25 & 90.61/89.47 & 90.53/88.96 & 73.61/73.85 & 73.50/74.36 & 73.02/74.96 & 73.21/75.77 & 73.71/74.54 & 73.70/74.62 & 73.24/75.12 & 73.90/\textbf{76.24} \\
            \cmidrule{2-14}
            & E & 89.57/89.39 & 89.43/88.89 & 89.57/89.39 & 89.43/88.89 & 73.73/75.67 & 73.68/74.68 & 73.73/75.67 & 73.68/74.68 & 73.95/74.90 & 74.01/75.10 & 73.95/74.90 & 74.01/75.10 \\
            \cmidrule{2-14}
            & F & 90.48/89.09 & 90.62/89.56 & 90.41/89.19 & 90.43/\textbf{89.94} & 72.86/74.18 & 73.82/\textbf{76.55} & 73.51/74.07 & 73.33/75.26 & 72.62/72.82 & 73.94/\textbf{76.66} & 73.32/74.08 & 74.50/75.84 \\
            \cmidrule{2-14}
            & G & 90.78/89.32 & 90.76/90.26 & 90.49/\textbf{89.82} & 89.47/89.60 & 74.33/73.35 & 73.63/73.59 & 74.05/75.05 & 73.22/74.34 & 74.57/73.78 & 73.31/74.02 & 74.21/75.13 & 73.87/74.80 \\
            \cmidrule{2-14}
            & H & \textbf{90.91}/88.88 & 90.45/88.98 & 90.29/88.93 & 89.99/89.14 & 74.43/73.68 & 73.62/74.56 & 73.59/74.06 & 73.36/74.71 & 74.48/73.96 & 73.65/74.62 & 73.90/74.12 & 73.42/75.04 \\
            \cmidrule{2-14}
            & I & 90.86/89.07 & 90.47/89.38 & 90.62/89.35 & 89.55/89.18 & 74.75/73.19 & 73.30/74.78 & 74.29/74.14 & 74.00/74.28 & 74.80/73.48 & 73.26/74.90 & 73.88/74.60 & 73.91/74.16 \\
            \cmidrule{2-14}
            & J & 89.43/89.23 & 89.65/89.30 & 89.53/88.99 & 89.89/89.43 & 73.75/\textbf{76.05} & 74.28/74.55 & 73.47/75.04 & \textbf{74.43}/74.95 & 74.05/\textbf{75.09} & \textbf{75.06}/74.97 & 73.52/\textbf{75.91} & \textbf{74.70}/75.53 \\
            \cmidrule{2-14}
            & K & 90.10/89.63 & 90.05/89.26 & 89.70/89.18 & 89.64/89.54 & 74.43/74.40 & 74.07/75.22 & \textbf{74.47}/75.89 & 74.38/74.97 & 74.44/74.80 & 74.23/75.42 & 74.30/75.81 & 74.02/74.67 \\
            \cmidrule{2-14}
            & L & 89.60/89.86 & 89.85/89.95 & 89.82/88.61 & 90.33/88.93 & 73.35/74.27 & \textbf{74.32}/75.42 & 73.68/\textbf{76.50} & 73.90/\textbf{76.15} & 73.16/74.55 & 74.04/75.52 & 73.66/75.08 & 73.52/76.02 \\
            \cmidrule{2-14}
            & M & 90.81/\textbf{90.27} & 90.07/89.85 & 90.01/89.30 & 89.75/89.73 & 74.21/74.59 & 74.29/74.85 & 73.96/74.94 & 74.32/74.87 & 74.03/74.04 & 74.36/74.68 & 73.72/75.10 & 73.95/75.59 \\
            \cmidrule{2-14}
            & N & 90.73/89.37 & 90.60/89.71 & 90.72/88.77 & 90.63/89.49 & 74.55/73.49 & 73.83/73.47 & 73.86/74.81 & 73.38/74.58 & 74.66/74.81 & 73.97/74.94 & 74.13/74.82 & 73.53/74.76 \\
            \cmidrule{2-14}
            & O & 90.90/89.94 & 90.50/89.35 & \textbf{90.90}/89.94 & 90.50/89.35 & 73.99/74.51 & 73.77/73.79 & 73.99/74.51 & 73.77/73.79 & 73.83/74.71 & 73.62/74.27 & 73.83/74.71 & 73.62/74.27 \\
            \cmidrule{2-14}
            & P & 89.72/89.80 & 90.31/90.08 & 89.13/89.19 & 90.30/89.92 & 73.37/75.03 & 73.87/75.02 & 73.57/75.24 & 74.41/75.44 & 73.48/74.79 & 74.66/75.18 & 73.54/74.84 & 73.52/75.79 \\
            \cmidrule{2-14}
            \cmidrule{2-14}
            & CD{\tnote{2}} & 87.42 & 88.93 & 87.76 & 88.10 & 73.09 & 75.04 & 74.15 & 75.35 & 73.64 & 74.94 & 74.38 & 75.44 \\
            \hline
            \parbox[t]{2mm}{\multirow{17}{*}{\rotatebox[origin=c]{90}{ReDAD}}} & A & 88.31/90.15 & 89.55/91.07 & 87.98/90.37 & 89.14/89.92 & 77.64/77.07 & 79.47/78.07 & 78.34/76.14 & \textbf{80.21}/78.17 & 77.34/77.21 & 79.26/77.83 & 78.44/76.40 & \textbf{80.13}/78.39 \\
            \cmidrule{2-14}
            & B & 87.82/90.57 & 89.11/90.52 & 87.66/88.83 & 88.64/87.11 & 77.74/77.56 & 78.65/78.24 & \textbf{78.61}/76.43 & 78.91/78.26 & 77.13/77.74 & 78.58/78.31 & 77.76/76.55 & 78.98/77.56 \\
            \cmidrule{2-14}
            & C & 88.30/89.64 & 89.21/87.01 & 88.11/88.79 & 89.17/89.87 & 77.14/76.70 & 79.67/77.89 & 78.19/76.41 & 79.32/77.59 & 77.08/77.05 & 79.35/77.92 & 77.82/76.62 & 79.26/77.78 \\
            \cmidrule{2-14}
            & D & 87.33/88.99 & 89.82/89.25 & 88.18/89.61 & 88.80/90.05 & 78.28/76.73 & 79.54/75.64 & 76.81/76.58 & 78.68/78.37 & 78.26/76.80 & 78.47/76.12 & 76.67/76.84 & 78.73/78.87 \\
            \cmidrule{2-14}
            & E & 88.03/88.63 & 89.37/90.91 & 88.03/88.63 & 89.37/90.91 & 77.83/77.45 & 77.54/78.30 & 77.83/77.45 & 77.54/78.30 & 77.75/77.34 & 77.44/78.41 & 77.75/77.34 & 77.44/78.41\\
            \cmidrule{2-14}
            & F & 87.71/89.54 & 88.45/89.45 & 87.87/90.11 & 88.54/90.99 & 77.59/76.50 & \textbf{79.74}/\textbf{79.23} & 76.94/76.75 & 79.68/77.94 & 77.35/76.33 & 79.19/\textbf{79.31} & 76.95/76.95 & 79.21/77.38 \\
            \cmidrule{2-14}
            & G & 88.17/90.06 & \textbf{89.83}/88.96 & 88.22/89.75 & \textbf{89.55}/90.15 & 77.83/77.64 & 79.39/78.61 & 78.13/77.04 & 79.09/77.91 & 77.50/77.69 & 79.12/78.76 & 77.88/77.56 & 78.74/77.73 \\
            \cmidrule{2-14}
            & H & 88.01/89.14 & 88.76/90.78 & 87.73/88.99 & 88.99/90.39 & 77.12/76.36 & 79.57/78.32 & 78.11/77.81 & 79.40/78.13 & 77.09/76.08 & 78.65/78.88 & 78.14/77.81 & 79.18/78.15 \\
            \cmidrule{2-14}
            & I & 87.56/88.64 & 88.05/89.67 & 87.86/90.14 & 89.45/90.13 & 77.77/76.29 & 79.23/77.70 & 78.40/76.08 & 78.99/78.42 & 77.11/76.56 & 79.05/78.31 & \textbf{78.49}/76.24 & 78.78/77.97\\
            \cmidrule{2-14}
            & J & 87.99/89.91 & 88.89/\textbf{91.12} & 87.79/90.50 & 89.06/89.05 & 77.69/77.48 & 78.40/78.92 & 77.14/78.35 & 78.40/77.55 & 77.71/77.57 & 78.57/78.87 & 76.59/77.92 & 78.18/77.43\\
            \cmidrule{2-14}
            & K & 88.36/\textbf{91.01} & 89.33/90.94 & 88.01/90.09 & 89.05/90.89 & 78.30/78.24 & 78.25/76.19 & 78.54/78.13 & 77.73/77.50 & 78.11/78.17 & 78.49/76.10 & 78.29/78.60 & 77.42/77.46\\
            \cmidrule{2-14}
            & L & 88.25/90.53 & 89.25/91.09 & 87.87/90.03 & 89.13/90.02 & \textbf{78.48}/77.87 & 78.94/77.67 & 77.88/77.59 & 77.91/77.97 & \textbf{78.52}/78.05 & 78.16/78.47 & 77.85/77.67 & 77.37/78.22\\
            \cmidrule{2-14}
            & M & \textbf{88.42}/90.07 & 89.57/90.80 & 88.12/90.52 & 89.40/90.46 & 77.00/77.31 & 78.85/78.42 & 78.02/78.50 & 77.12/77.66 & 76.62/77.20 & \textbf{79.66}/77.85 & 78.02/78.38 & 77.08/77.66\\
            \cmidrule{2-14}
            & N & 87.98/90.71 & 88.94/90.82 & 88.08/90.43 & 89.47/90.68 & 78.21/77.97 & 78.92/78.24 & 78.03/77.63 & 78.78/78.06 & 78.07/77.30 & 77.91/78.39 & 77.74/77.73 & 78.60/78.10\\
            \cmidrule{2-14}
            & O & 88.27/90.59 & 87.71/91.06 & 88.27/\textbf{90.59} & 87.71/\textbf{91.06} & 77.08/\textbf{79.02} & 78.96/78.78 & 77.08/\textbf{79.02} & 78.96/\textbf{78.78} & 76.78/\textbf{78.95} & 79.03/78.97 & 76.78/\textbf{78.95} & 79.03/\textbf{78.97}\\
            \cmidrule{2-14}
            & P & 88.02/89.41 & 88.86/89.93 & \textbf{88.33}/90.26 & 89.51/87.85 & 78.43/77.45 & 79.38/76.67 & 78.44/77.19 & 79.12/77.70 & 78.37/76.25 & 79.44/77.27 & 78.04/77.31 & 78.91/77.76\\
            \cmidrule{2-14}
            \cmidrule{2-14}
            & CD{\tnote{2}} & 88.40 & 89.33 & 89.01 & 89.16 & 73.69 & 74.29 & 72.67 & 72.81 & 74.13 & 74.82 & 72.91 & 73.76 \\
            \hline
        \end{tabular}
    \begin{tablenotes}
      \item [1] Type of Relation Representation.
      \item [2] Cross-disease experiments utilizing the relation representation $R_A$: Training on ReDReS, evaluation on ReDAD, and vice versa.
    \end{tablenotes}
  \end{threeparttable}}
\end{table*}

\subsection{Experiments with Additional Language Models}

Assessing the impact of different language models on the performance of the LaMReDA and LaMEL models, we conduct experiments using additional models to replace the BioMedBERT model (Fig. \ref{fig:model_architecture}). Specifically, we evaluate the effect of BioLinkBERT \cite{yasunaga-etal-2022-linkbert} (base model) and BioGPT \cite{luo2022biogpt}, both of which are pre-trained on PubMed abstracts. BioLinkBERT is selected to determine whether leveraging the links between documents in the pre-training corpus benefits the relation detection task of our study. This model uses citations between PubMed articles to model document connections, which may enhance its ability to capture semantic relations. BioGPT, as a GPT-based \cite{radford2018improving, radford2019language} transformer decoder \cite{vaswani2017attention} representative, is incorporated to explore the generative model's capabilities in capturing semantic relations between entities. \cite{yasunaga-etal-2022-linkbert} state that the pre-training data of BioLinkBERT and BioMedBERT are identical\footnote{\url{https://pubmed.ncbi.nlm.nih.gov}. Papers published before Feb. 2020.}, except that they also use the citations between PubMed articles. \cite{luo2022biogpt} clarify that BioGPT's pre-training includes PubMed abstracts updated before 2021. Hence, all language models in our study are pre-trained on identical or nearly identical datasets, making the comparison between them reliable.
\par
We conduct experiments using a variety of representations, specifically testing the $R_A$ \eqref{eq:r_a}, $R_D$ \eqref{eq:r_d}, $R_E$ \eqref{eq:r_e}, $R_F$ \eqref{eq:r_f}, $R_G$ \eqref{eq:r_g}, and $R_O$ \eqref{eq:r_o} relation representations of LaMReDA, as well as the $E_A$ \eqref{eq:e_a}, $E_D$ \eqref{eq:e_d}, $E_E$ \eqref{eq:e_e}, and $E_H$ \eqref{eq:e_h}) entity representations of LaMEL. The experimental setup follows the same procedure outlined in Subsection \ref{subsec:experimental_setup} (Table \ref{tab:experimental_parameters}), and the experiments are performed on both the original split (Table \ref{tab:datasets_statistics}) and in a 5-fold cross-validation setting.

\begin{table*}[!ht]
  \caption{\label{tab:lamreda_diff_lms} LaMReDA Results (\%) in binary and multi-class setup utilizing BiomedBERT-base, BioLinkBERT-base, and BioGPT as backbone language model: Each cell shows the average F1-score from 10 runs (original test set) and from 5-fold cross-validation setup.}
  \resizebox{\textwidth}{!}{
  \begin{threeparttable}
        \begin{tabular}{ccccc|ccc|ccc}
            \hline
            \multirow{3}{*}{\textbf{Data}} & \multirow{3}{*}{\textbf{Type}\tnote{1}} & \multicolumn{3}{c}{\textbf{Binary setup}} & \multicolumn{6}{|c}{\textbf{Multi-class setup}}\\
            \cmidrule{6-11}
            & & & & & \multicolumn{3}{c}{\textbf{Micro Evaluation}} & \multicolumn{3}{|c}{\textbf{Macro Evaluation}} \\
            \cmidrule{3-11}
            & & \textbf{BiomedBERT\tnote{2}} & \textbf{BioLinkBERT\tnote{2}} & \textbf{BioGPT} & \textbf{BiomedBERT\tnote{2}} & \textbf{BioLinkBERT\tnote{2}} & \textbf{BioGPT} & \textbf{BiomedBERT\tnote{2}} & \textbf{BioLinkBERT\tnote{2}} & \textbf{BioGPT}\\
            \hline
            \parbox[t]{2mm}{\multirow{6}{*}{\rotatebox[origin=c]{90}{ReDReS}}} & A & 90.72/89.95 & \textbf{91.15}/89.90 & \textit{90.15}/\textit{88.83} & \textbf{74.49}/73.91 & 74.06/\textit{75.13} & 72.22/72.65 & 74.52/74.50 & 74.11/75.09 & 71.80/72.95 \\
            \cmidrule{2-11}
            & D & 90.55/\textit{89.29} & 90.35/90.12 & 89.76/88.08 & 73.61/73.85 & 74.12/75.19 & 71.42/73.25 & 73.71/74.54 & 74.12/75.54 & 72.48/73.63 \\
            \cmidrule{2-11}
            & E & 89.57/89.39 & 89.98/\textbf{90.28} & 89.49/87.19 & 73.73/\textbf{75.67} & 73.22/74.40 & 70.17/70.98 & 73.95/\textit{74.90} & 73.38/74.97 & 69.97/70.69 \\
            \cmidrule{2-11}
            & F & 90.48/89.09 & 90.72/89.84 & 89.68/88.35 & 72.86/74.18 & \textit{74.22}/75.12 & 71.36/\textit{73.57} & 72.62/72.82 & 74.13/75.10 & 72.00/\textit{73.82} \\
            \cmidrule{2-11}
            & G & 90.78/89.32 & 90.57/89.57 & 89.93/88.41 & 74.33/73.35 & 74.00/75.65 & \textit{72.68}/73.14 & \textbf{74.57}/73.78 & \textit{74.32}/\textbf{75.94} & \textit{72.48}/73.43 \\
            \cmidrule{2-11}
            & O & \textit{90.90}/89.94 & 90.75/90.12 & 89.51/87.82 & 73.99/74.51 & 74.00/73.65 & 66.48/68.67 & 73.83/74.71 & 73.69/74.03 & 67.43/67.75 \\
            \hline
            \parbox[t]{2mm}{\multirow{6}{*}{\rotatebox[origin=c]{90}{ReDAD}}} & A & \textit{88.31}/90.15 & 88.73/89.18 & 87.49/89.19 & 77.64/77.07 & 78.40/76.38 & 76.04/75.96 & 77.34/77.21 & 77.68/76.66 & 76.45/75.62 \\
            \cmidrule{2-11}
            & D & 87.33/88.99 & \textbf{88.74}/89.72 & 87.56/\textit{89.63} & \textit{78.28}/76.73 & 78.27/77.03 & \textit{76.49}/\textit{76.02} & \textbf{78.26}/76.80 & 77.89/77.18 & 76.25/76.04 \\
            \cmidrule{2-11}
            & E & 88.03/88.63 & 88.43/\textit{90.48} & \textit{87.71}/87.86 & 77.83/77.45 & \textbf{78.96}/\textit{78.02} & 75.35/73.65 & 77.75/77.34 & \textit{78.00}/\textit{78.10} & 75.36/73.28 \\
            \cmidrule{2-11}
            & F & 87.71/89.54 & 88.62/89.86 & 87.63/88.94 & 77.59/76.50 & 77.43/77.25 & 76.34/75.93 & 77.35/76.33 & 77.48/77.45 & 75.47/\textit{76.10} \\
            \cmidrule{2-11}
            & G & 88.17/90.06 & 88.72/90.38 & 87.55/88.85 & 77.83/77.64 & 77.76/76.99 & 76.35/75.71 & 77.50/77.69 & 77.33/76.41 & \textit{76.61}/75.73 \\
            \cmidrule{2-11}
            & O & 88.27/\textbf{90.59} & 88.58/89.91 & 86.19/87.74 & 77.08/\textbf{79.02} & 76.89/74.85 & 71.69/71.32 & 76.78/\textbf{78.95} & 76.62/74.93 & 70.57/70.75 \\
            \hline
        \end{tabular}
    \begin{tablenotes}
      \item [1] Type of Relation Representation.
      \item [2] Base version.
    \end{tablenotes}
  \end{threeparttable}}
\end{table*}

\begin{table}[!ht]
  \caption{\label{tab:LaMEL_results}LaMEL Results (\%) in binary setup (BiomedBERT $\square$: base, $\blacksquare$:large): Each cell (unless cross-disease experiments) shows the average F1-score from 10 runs (original test set) and from 5-fold cross-validation setup.}
  \resizebox{\columnwidth}{!}{
  \begin{threeparttable}
        \begin{tabular}{cc|c|c|c}
            \hline
            \multirow{3}{*}{\textbf{Type}\tnote{1}} & \multicolumn{2}{c}{\textbf{ReDReS}} & \multicolumn{2}{|c}{\textbf{ReDAD}}\\
            \cmidrule{2-5}
            & \textbf{F\textsubscript{1}}\textsuperscript{$\square$} & \textbf{F\textsubscript{1}}\textsuperscript{$\blacksquare$} & \textbf{F\textsubscript{1}}\textsuperscript{$\square$} & \textbf{F\textsubscript{1}}\textsuperscript{$\blacksquare$} \\
            \hline
            A & 90.25/89.43 & 90.88/90.01 & 86.73/88.75  & 88.90/90.17\\
            \hline
            B & 90.29/89.01 & 90.73/89.41 & 86.89/89.29 & 88.22/89.15 \\
            \hline
            C & 90.51/89.44 & 90.71/89.67 & \textbf{87.49}/\textbf{90.02} & 88.57/90.65 \\
            \hline
            D & 90.47/88.90 & \textbf{91.03}/90.07 & 86.29/88.88 & 88.22/90.64 \\
            \hline
            E & \textbf{90.61}/89.10 & 90.54/89.55 & 86.03/88.96 & 88.74/90.35\\
            \hline
            F & 90.48/89.37 & 90.88/\textbf{90.29} & 87.27/89.18 & \textbf{89.44}/90.57\\
            \hline
            G & 90.32/\textbf{89.71} & 90.35/89.43 & 86.97/89.46 & 89.12/\textbf{91.25}\\
            \hline
            H & 89.68/89.24 & 90.13/89.29 & 87.29/89.91 & 88.77/90.67\\
            \hline
            \hline
            CD{\tnote{2}} & 86.20 & 89.14 & 88.92 & 88.56\\
            \hline
        \end{tabular}
    \begin{tablenotes}
      \item [1] Type of Relation Representation.
      \item [2] Cross-disease experiments utilizing the entity representation $E_A$: Training on ReDReS, evaluation on ReDAD, and vice versa.
    \end{tablenotes}
  \end{threeparttable}}
\end{table}

\begin{table}[!ht]
  \caption{\label{tab:lamel_diff_lms} LaMEL Results (\%) in binary setup utilizing BiomedBERT-base, BioLinkBERT-base, and BioGPT as backbone language model: Each cell shows the average F1-score from 10 runs (original test set) and from 5-fold cross-validation setup.}
  \resizebox{\columnwidth}{!}{
  \begin{threeparttable}
        \begin{tabular}{cccc|ccc}
            \hline
            \multirow{3}{*}{\textbf{Type}\tnote{1}} & \multicolumn{3}{c}{\textbf{ReDReS}} & \multicolumn{3}{|c}{\textbf{ReDAD}}\\
            \cmidrule{2-7}
            & \textbf{BiomedBERT\tnote{2}} & \textbf{BioLinkBERT\tnote{2}} & \textbf{BioGPT} & \textbf{BiomedBERT\tnote{2}} & \textbf{BioLinkBERT\tnote{2}} & \textbf{BioGPT} \\
            \hline
            A & 90.25/\textit{89.43} & 90.68/89.02 & 89.43/86.64 & 86.73/88.75 & \textbf{87.71}/89.09 & 86.14/\textit{88.54} \\
            \hline
            D & 90.47/88.90 & 90.35/89.30 & 88.52/\textit{88.17} & 86.29/88.88 & 87.35/88.70 & 85.28/87.85 \\
            \hline
            E & \textit{90.61}/89.10 & 90.25/88.08 & \textit{89.48}/87.24 & 86.03/88.96 & 87.05/\textit{89.36} & \textit{87.42}/87.96 \\
            \hline
            H & 89.68/89.24 & \textbf{90.73}/\textbf{89.48} & 89.21/87.73 & \textit{87.29}/\textbf{89.91} & 87.56/88.83 & 85.55/88.23 \\
            \hline
            \hline
        \end{tabular}
    \begin{tablenotes}
      \item [1] Type of Relation Representation.
      \item [2] Base version.
    \end{tablenotes}
  \end{threeparttable}}
  \vspace{-4mm}
\end{table}

\section{Results}
\label{sec:results}

Tables \ref{tab:LaMReDA_LaMReDM_results} and \ref{tab:LaMEL_results} report the F1-scores for the LaMReDA, LaMReDM, and LaMEL models on the ReDReS and ReDAD datasets. Each cell (except for the cross-disease experiments) displays two values: the average F1-score from 10 runs on the original test set (Table \ref{tab:datasets_statistics}) and the average F1-score from a 5-fold cross-validation. The models perform well across all relation (A-P) and entity (A-H) representations, showing their ability to learn meaningful representations for the semantic relation task regardless of initial token selection. However, we observe patterns regarding relation representations. In the binary setup, the relation representation $R_G$ \eqref{eq:r_g} yields strong results for both datasets, suggesting that including the \textit{[CLS]} token representation might be beneficial. In the multi-class setup, relation representations $R_L$ \eqref{eq:r_l}, $R_J$ \eqref{eq:r_j}, and $R_O$ \eqref{eq:r_o} are effective for both datasets, indicating that the surrounding context is crucial for the more complex task, as $R_L$ and $R_J$ include the averaged pooled representation of intermediate tokens between entities, and $R_O$ leverages the context vector \cite{zhou2021document}. The intra-model comparison reveals that over-parameterization tends to be useful. Using BiomedBERT large generally results in better performance than the base alternative. The BiomedBERT base shows superior performance mainly only in experiments using the original splits of ReDReS (Table \ref{tab:datasets_statistics}). LaMEL is highly competitive with LaMReDA and LaMReDM, indicating that learning entity embedding spaces optimized for relation detection is promising. LaMEL achieves the highest performance in the 5-fold setup of ReDAD and the original setup of ReDReS, with F1-scores of 91.03\% and 91.25\% respectively.

The inter-model comparison across the same relation representations indicates that the aggregation function does not significantly impact the relation detection task. Neither LaMReDA (element-wise addition) nor LaMReDM (element-wise multiplication) show a clear advantage over the other. This suggests that the transformer layers of BiomedBERT and the projection layer $l()$ preceding the aggregation are effectively trained in both models to encode the essential information for relation detection, regardless of the aggregation function used. The cross-disease experiments underscore the robustness of the models in both binary and multi-class setups. This robustness supports transfer learning \cite{zhuang2020comprehensive} in semantic relation detection, extending to other diseases, highlighting the potential for broader applications and research endeavors in knowledge discovery.

\noindent \textbf{Human Performance.} To assess and compare the human performance, two additional experts identify the relation type in a random sample of 300 instances from the test set of each dataset (Table \ref{tab:datasets_statistics}). The evaluation ground truth is based on the original test set labels. In the binary setup, the average F1-score ranges from 92.14 for ReDReS to 91.87 for ReDAD. The LaMReDA, LaMReDM, and LaMEL models achieve performance comparable to those of human experts, indicating a high ability to detect semantic relations. Multi-class macro F1-scores range from 85.23 (micro: 85.45) to 85.76 (micro: 85.87) for ReDReS and ReDAD, respectively. Compared with human experts, all models show a performance gap, highlighting that identifying more complex aspects of semantic relations is a challenging task.

\noindent \textbf{Baseline performance - lower bound.} Labels are randomly assigned labels based on the class distribution of the training data (Table \ref{tab:datasets_statistics}). In the binary setup, the baseline achieves F1-scores of 54\% (ReDReS) and 53.16\% (ReDAD). For the multi-class setup, the macro F1-scores range from 32.05\% to 32.43\%, stressing the task's difficulty, particularly for distinguishing various semantic relations (multi-class). Appendix \ref{sec:baseline_performance_appendix} provides additional information.

The results presented in Tables \ref{tab:lamreda_diff_lms} and \ref{tab:lamel_diff_lms} demonstrate the F1-scores of the LaMReDA and LaMEL models on the ReDReS and ReDAD datasets, using different language models. The experiments show that both LaMReDA and LaMEL perform consistently well across various relations ($R_A$, $R_D$, $R_E$, $R_F$, $R_G$, $R_O$) and entities ($E_A$, $E_D$, $E_E$, $E_H$) representations, utilizing the different language models. This indicates the models’ ability to learn meaningful representations for the semantic relation detection task, independent of the initial token selection.

An intra-lm comparison reveals that the use of relation representation $R_O$ results in lower performance across both datasets when BioGPT is employed, particularly in the multi-class setup (Table \ref{tab:lamreda_diff_lms}). This suggests that the context vector extracted from the generative model is less informative for this task. Conversely, when the context vector is derived from BiomedBERT and BioLinkBERT, the models show strong performance, showing the best results on ReDAD in the 5-fold cross-validation setup (Table \ref{tab:lamreda_diff_lms}).

The inter-language model comparison indicates that BiomedBERT and BioLinkBERT perform similarly, suggesting that leveraging document links during pre-training, as BioLinkBERT does, does not offer a significant advantage for our task. In contrast, the use of BioGPT leads to lower performance, particularly in the multi-class setup, emphasizing the superiority of encoder-based language models such as BiomedBERT and BioLinkBERT over decoder-based BioGPT for the relation detection task. Despite BioGPT’s larger size, with 347 million parameters \cite{luo2022biogpt} compared with the 110 million parameters of BiomedBERT and BioLinkBERT \cite{vaswani2017attention, yasunaga-etal-2022-linkbert}, the generative approach appears less effective for this particular semantic relation task, further underscoring the advantages of encoder-focused models in this context.

\begin{figure}[!ht]
  \centering
  \includegraphics[width=\columnwidth]{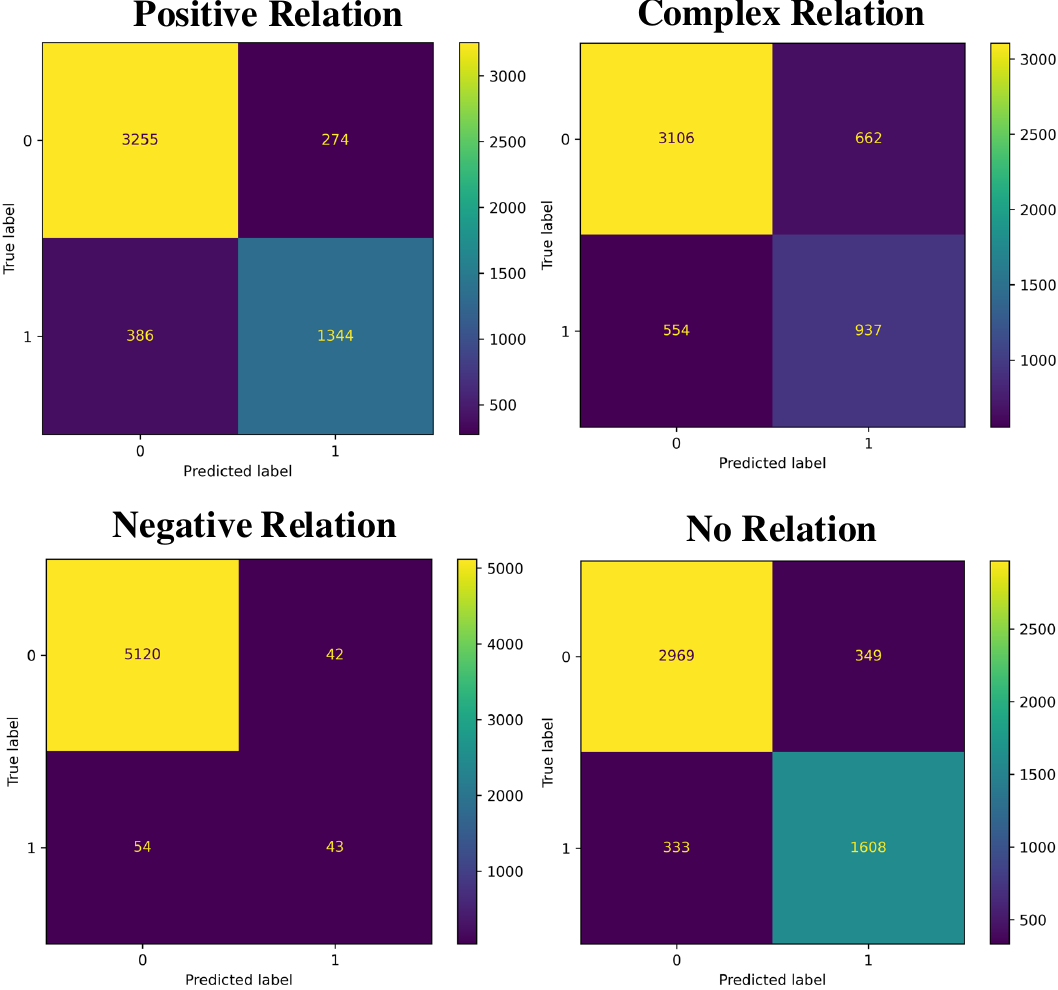}
  \caption{\label{fig:cm_biomedbert_base_redres}ReDReS Confusion Matrices for each relation class using the best-performing model on the ReDReS dataset based on the macro evaluation (Table \ref{tab:LaMReDA_LaMReDM_results}), with BiomedBERT-base as the backbone language model.}
\end{figure}

\begin{figure}[!ht]
  \centering
  \includegraphics[width=\columnwidth]{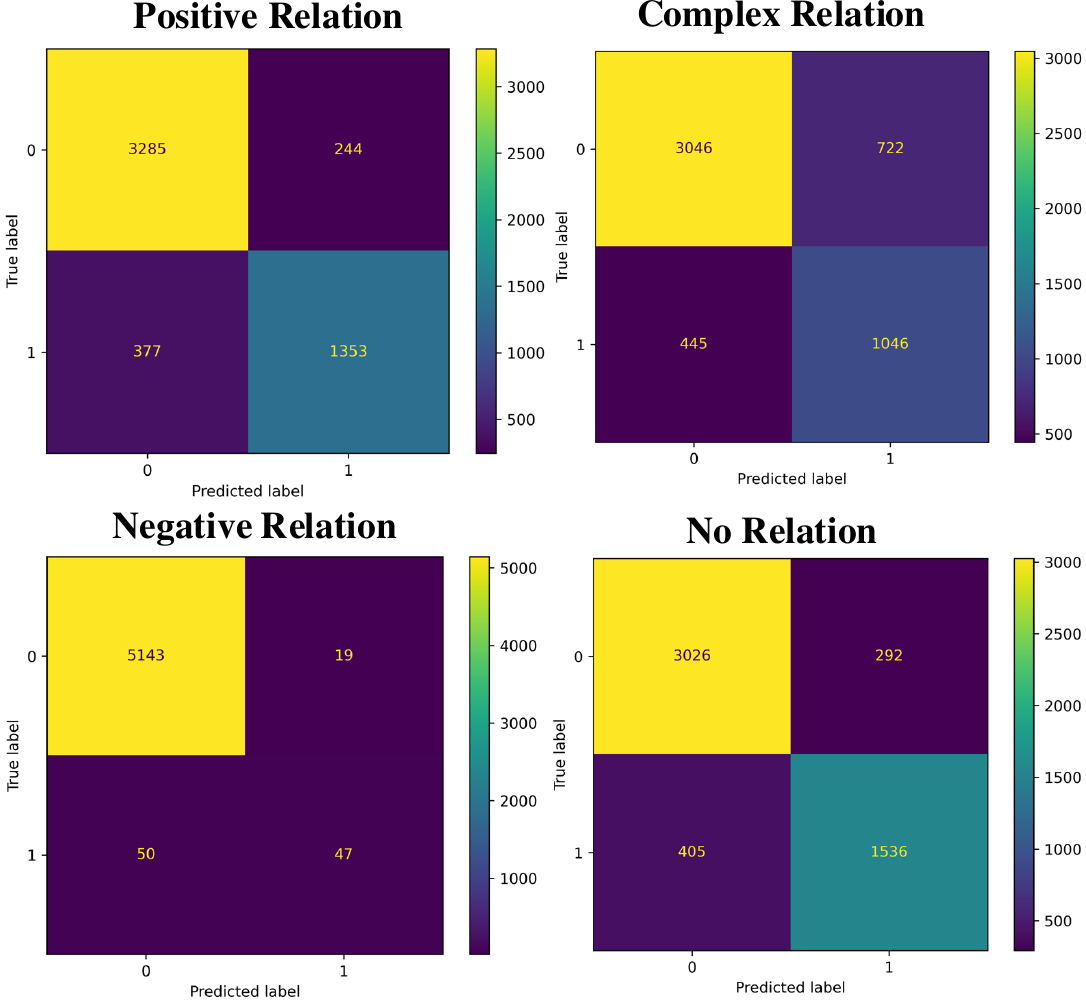}
  \caption{\label{fig:cm_biomedbert_large_redres}ReDReS Confusion Matrices for each relation class using the best-performing model on the ReDReS dataset based on the macro evaluation (Table \ref{tab:LaMReDA_LaMReDM_results}), with BiomedBERT-large as the backbone language model.}
\end{figure}

\subsection{Error Analysis}

\begin{figure}[!h]
  \centering
  \includegraphics[width=\columnwidth]{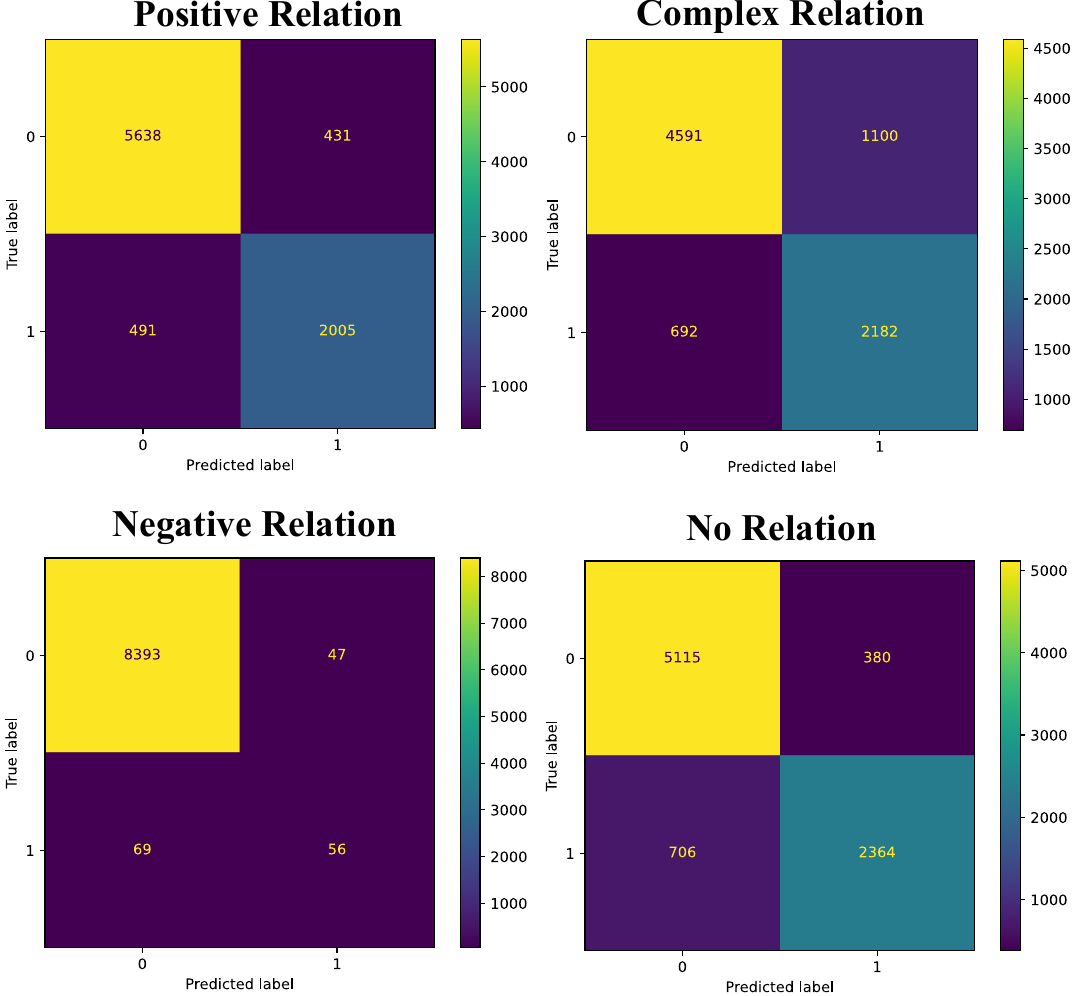}
  \caption{\label{fig:cm_biomedbert_base_redad}ReDAD Confusion Matrices for each relation class using the best-performing model on the ReDAD dataset based on the macro evaluation (Table \ref{tab:LaMReDA_LaMReDM_results}), with BiomedBERT-base as the backbone language model.}
\end{figure}

\begin{figure}[!h]
  \centering
  \includegraphics[width=\columnwidth]{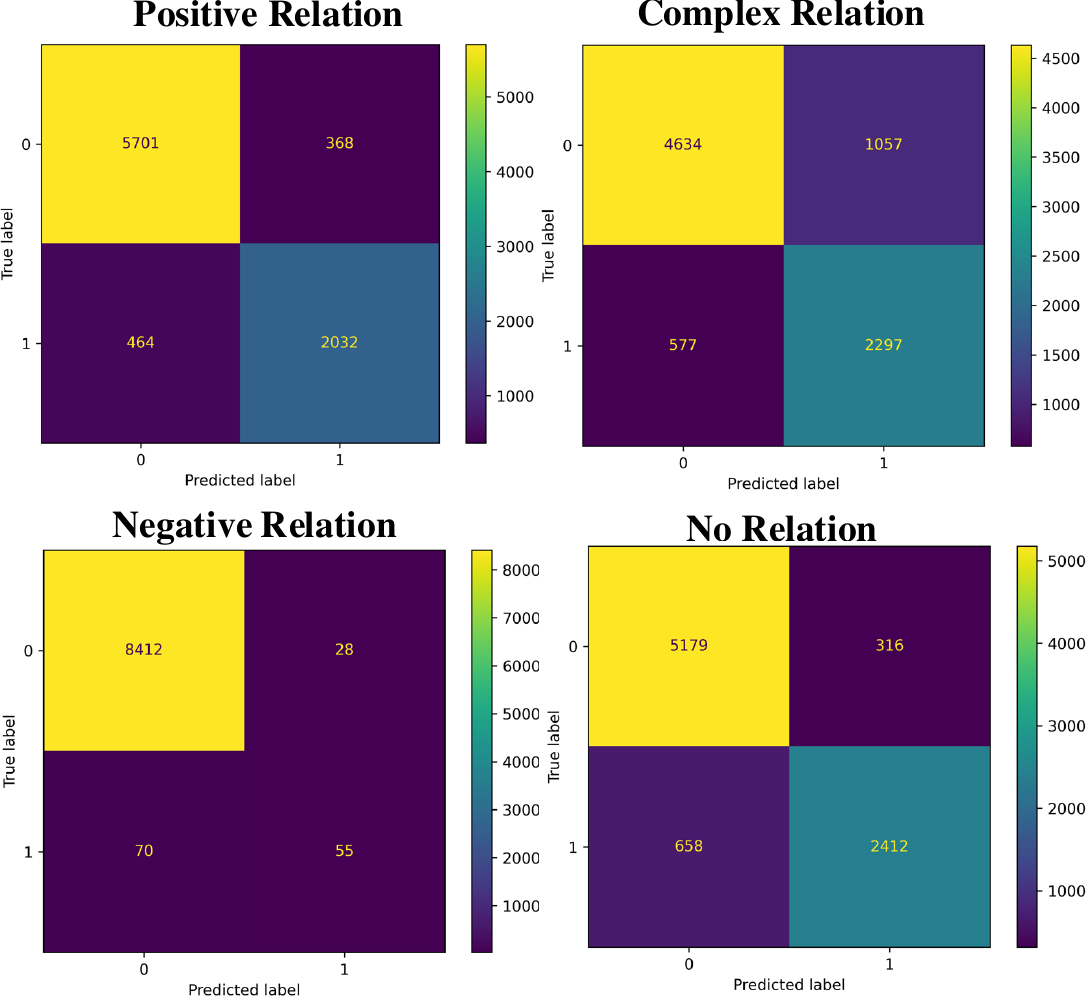}
  \caption{\label{fig:cm_biomedbert_large_redad}ReDAD Confusion Matrices for each relation class using the best-performing model on the ReDAD dataset based on the macro evaluation (Table \ref{tab:LaMReDA_LaMReDM_results}), with BiomedBERT-large as the backbone language model.}
\end{figure}

Whereas the macro and micro evaluations in the multi-class setup indicate strong overall performance across different relation types, we conduct an error analysis to provide additional qualitative insights. Using the best-performing models on the ReDReS and ReDAD datasets based on the macro evaluation (Table \ref{tab:LaMReDA_LaMReDM_results}), we extract predictions from the 5-fold cross-validation setup. Confusion matrices (tables that are used to analyze the classification performance, by displaying the number of true positives, true negatives, false positives, and false negatives) for each class are generated to present a comprehensive view of the misclassified instances (Fig. \ref{fig:cm_biomedbert_base_redres}, \ref{fig:cm_biomedbert_large_redres}, \ref{fig:cm_biomedbert_base_redad}, \ref{fig:cm_biomedbert_large_redad}). For example, the confusion matrix for the \textit{positive} class is presented in the top-left side of Fig. \ref{fig:cm_biomedbert_base_redres}, where the number of true positives, true negatives, false positives, and false negatives are 1344, 3255, 274 and 386, respectively. 

We analyze the false negatives for each relation class to gain insights into the misclassification rates. In the ReDReS dataset, the majority of misclassified instances of the \textit{positive} relation (324 and 331 instances using BiomedBERT-base and BiomedBERT-large, respectively) are labeled as \textit{complex} relations (Table \ref{tab:fn_analysis}). This is intuitively reasonable, as the model recognizes that the entities are semantically related but fails to predict the correct relation type. Conversely, \textit{complex} relation instances are often mislabeled as \textit{no relation} (Table \ref{tab:fn_analysis}), highlighting the challenge of detecting nuanced semantic relations that require deeper reasoning. For the \textit{no relation} class (Table \ref{tab:fn_analysis}), most false negatives (364 and 303 instances using BiomedBERT-base and BiomedBERT-large, respectively) are also labeled as \textit{complex} relations, indicating a model bias towards predicting associations between co-occurring entities. Similar patterns are observed in the ReDAD dataset regarding the \textit{positive}, \textit{complex}, and \textit{no relation} classes. The under-represented \textit{negative} class has the most misclassified cases labeled as \textit{complex} or \textit{positive} relations in ReDReS and ReDAD (Table \ref{tab:fn_analysis}), respectively, suggesting that while the model detects the semantic connection, it tends to predict one of the two more represented classes in the datasets.

Different strategies can be applied during the training phase to improve model performance to address the class imbalance in the datasets. In the multi-class setup, the classes such as \textit{positive}, \textit{complex}, and \textit{no relation} are over-represented. We can handle this imbalance by undersampling these over-represented classes, reducing their instances to approximate the frequency of the under-represented \textit{negative} class. This approach ensures that the model does not become biased toward the majority classes, encouraging it to learn from a more balanced distribution of samples. However, due to the limited number of under-representative class instances, oversampling of \textit{negative} class by incorporating additional samples from the distantly supervised datasets may be a more effective strategy. By increasing the number of instances of the \textit{negative} class, the model may learn its distinct characteristics, thereby enhancing the classification performance. Another approach to managing class imbalance is to adjust the loss function by increasing the weight of the under-represented class. By assigning a higher rescaling weight to this class during loss calculation, the model is penalized more heavily for misclassifying instances from the \textit{negative} class. This encourages the model to focus on improving its performance on this less frequent class, balancing the impact of each class on the overall learning process.

\begin{table}[!t]
  \caption{\label{tab:fn_analysis}False negative analysis for the \textit{Positive}, \textit{Complex}, \textit{Negative}, and \textit{No Relation} relation class.}
  \resizebox{\columnwidth}{!}{
        \begin{tabular}{c|c|cccc|cccc}
            \hline
            \multirow{2}{*}{\textbf{Data}} & \multirow{2}{*}{\textbf{Class}} & \multicolumn{4}{c}{\textbf{BiomedBERT-base}} & \multicolumn{4}{|c}{\textbf{BiomedBERT-large}} \\
            \cmidrule{3-10}
            & & \textbf{Positive} & \textbf{Complex} & \textbf{Negative} & \textbf{No Relation} & \textbf{Positive} & \textbf{Complex} & \textbf{Negative} & \textbf{No Relation} \\
            \hline
            \multirow{4}{*}{ReDReS} & Positive & - & 331 & 14 & 41 & - & 324 & 11 & 42 \\
            & Complex & 221 & - & 26 & 307 & 188 & - & 8 & 249 \\
            & Negative & 25 & 28 & - & 1 & 15 & 34 & - & 1 \\
            & No Relation & 28 & 303 & 2 & - & 41 & 364 & 0 & - \\
            \hline
            \multirow{4}{*}{ReDAD} & Positive & - & 424 & 27 & 40 & - & 425 & 21 & 18 \\
            & Complex & 335 & - & 19 & 338 & 274 & - & 6 & 297 \\
            & Negative & 44 & 23 & - & 2 & 44 & 25 & - & 1 \\
            & No Relation & 52 & 653 & 1 & - & 50 & 607 & 1 & - \\
            \hline
        \end{tabular}}
    \vspace{-2mm}
\end{table}

\section{Distantly Supervised Datasets}
\label{sec: distantly_supervised_datasets}

ReDReS and ReDAD include gold annotations for a small fraction of the extracted sentences. The pre-processed text consists of 28,622 and 1,301,429 additional sentences related to RS and AD respectively, without annotations of the semantic relation between the detected entities. Observing that the supervised models achieve performance levels comparable to those of human experts, we leverage the best-performing models to generate silver labels for the unannotated instances. For the binary setup, we employ:

\begin{itemize}
    \item LaMReDA (BiomedBERT large) with the relation representation $R_A$ \eqref{eq:r_a} for the RS corpus.
    \item LaMReDA (BiomedBERT large) with the relation representation $R_J$ \eqref{eq:r_j} for the AD corpus.
\end{itemize}

\noindent For the multi-class setup, we use:

\begin{itemize}
    \item LaMReDA (BiomedBERT large) with the relation representation $R_F$ \eqref{eq:r_f} for the RS corpus.
    \item LaMReDA (BiomedBERT large) with the relation representation $R_F$ \eqref{eq:r_f} for the AD corpus.
\end{itemize}

We stress that the selection of the models relies on the performance in the 5-fold cross-validation setup to avoid choosing based on the model performance on the original test set. Each model is trained 10 times with different seeds using the original splits (Table \ref{tab:datasets_statistics}) of ReDReS and ReDAD. The best model weights are saved based on the performance on the development set. Every trained model provides predictions for the unannotated instances and the final silver labels are extracted through majority voting.  
The \textbf{Di}stantly \textbf{S}upervised \textbf{Re}lation \textbf{D}etection dataset for \textbf{Re}tt \textbf{S}yndrome (DiSReDReS) contains 304,008 instances with 8,611 unique CUIs and 80 semantic types (Tables \ref{tab:distant_datasets_statistics} and \ref{tab:distant_datasets_distribution}). The \textbf{Di}stantly \textbf{S}upervised \textbf{Re}lation \textbf{D}etection dataset for \textbf{A}lzheimer's \textbf{D}isease (DiSReDAD) comprises 13,608,175 instances with 53,750 unique CUIs and 82 semantic types (Tables \ref{tab:distant_datasets_statistics} and \ref{tab:distant_datasets_distribution}). As noisy labeling is inevitable in distantly supervised data and imposes challenges for knowledge extraction scenarios, the two extensive datasets can promote weakly supervised learning.

\noindent
\textbf{Weakly Supervised Setup.} The task formulation remains the same as that described in Section \ref{sec:models}. The train sets of ReDReS and ReDAD are replaced by DiSReDReS and DiSReDAD, respectively. The development and test sets are identical (Table \ref{tab:datasets_statistics}). To provide a benchmark, we train the LaMReDA (BiomedBERT base) with the relation representation $R_A$ \eqref{eq:r_a} for 10 epochs utilizing the ADAM optimizer with learning rate 10\textsuperscript{-5}. The batch is set to 32. The experiments are repeated 10 times with different seeds (42, 3, 7, 21, 77, 24, 69, 96, 44, 11) and the best scores are retained based on the performance on the development set.

\begin{table}[!b]
  \caption{\label{tab:distant_datasets_statistics}DiSReDReS and DiSReDAD: Statistics and performance (\%)}
  \resizebox{\columnwidth}{!}{
  \begin{threeparttable}
        \begin{tabular}{cccccccc}
            \hline
            \multirow{3}{*}{\textbf{Data}} & \multirow{3}{*}{\textbf{Sentences}} & \multirow{3}{*}{\textbf{Instances}} & \multirow{3}{*}{\textbf{CUIs}\tnote{1}} & \multirow{3}{*}{\textbf{S.T.}\tnote{1}} & \multicolumn{3}{c}{\textbf{Benchmark}\tnote{3}}\\
            & & & & & \textbf{Binary} & \multicolumn{2}{c}{\textbf{Multi-Class}}\\
            & & & & & & \textbf{Micro} & \textbf{Macro}\\
            \hline
            DiSReDReS & 28,622 & 304,008 & 8,611 & 80 & 91.22\textpm{0.3} & 74.54\textpm{0.8} & 74.65\textpm{0.9}\\
            DiSReDAD & 1,301,429 & 13,608,175 & 53,750 & 82 & 89.40\textpm{0.3} & 80.79\textpm{0.3} & 80.81\textpm{0.4}\\
            \hline
        \end{tabular}
    \begin{tablenotes}
      \item [1] The total number of unique CUIs.
      \item [2] Semantic Types.
      \item [3] The benchmark performance (F1-score \%) in the weakly supervised setup.
    \end{tablenotes}
  \end{threeparttable}}
\end{table}

\begin{table}[!b]
   \caption{\label{tab:distant_datasets_distribution}DiSReDReS and DiSReDAD: Label Distribution}
    \centering
    \resizebox{\columnwidth}{!}{
        \begin{tabular}{ccccc}
        \hline
        \multirow{2}{*}{\textbf{Dataset}} & \multicolumn{4}{c}{\textbf{Labels - Type of relation}} \\
        & \textbf{Positive} & \textbf{Complex} & \textbf{Negative} & \textbf{No Relation} \\
        \hline
        DiSReDReS & 97,099 (31.9\%) & 105,861 (34.8\%) & 3,242 (1.1\%) & 97,806 (32.2\%)\\
        \hline
        DiSReDAD & 4,468,110 (32.8\%) & 5,755,884 (42.3\%) & 120,267 (0.9\%) & 3,263,914 (24.0\%)\\
        \hline
    \end{tabular}}
\end{table}

In the supervised setup (Table \ref{tab:datasets_statistics}), LaMReDA (BiomedBERT base) with the $R_A$ representation achieves 90.72\% and 88.31\% F1-score in the binary setup on ReDReS and ReDAD, respectively (Table \ref{tab:LaMReDA_LaMReDM_results}). The multi-class macro F1-scores range from 74.52\% (micro: 74.49\%) to 77.34\% (micro: 77.64\%) for ReDReS and ReDAD, accordingly (Table \ref{tab:LaMReDA_LaMReDM_results}). Table \ref{tab:distant_datasets_statistics} presents the benchmark performance in the binary and multi-class setup for both datasets.  Notably, the performance is improved in the weakly supervised setup, indicating the robustness of LaMReDA, when trained with noisy data, and highlighting the quality of the silver labels of DiSReDReS and DiSReDAD.

\section{Probing}
\label{sec:probing}

This study probes BiomedBERT's ability to capture semantic relations between entities. We explore different transformer layer representations and attention scores per layer and attention head. The averaged pooled entity representations are extracted from each layer, followed by training a linear classification layer. We test relation representations $R_D$ \eqref{eq:r_d}, $R_O$ \eqref{eq:r_o}, and $R_P$ \eqref{eq:r_p} of LaMReDA and LaMReDM to assess the impact of the context vector. Out-of-the-box representations are evaluated without the projection linear layer $l()$. We also extract the average attention scores of tokens for each entity towards the other across each layer and head, concatenating them into a feature vector for training a linear classification layer. Following \cite{chizhikova-etal-2022-attention}, we also train the classification layer using the average attention scores between the two entities across all layers.

\begin{figure}[!t]
  \centering
  \includegraphics[width=\columnwidth]{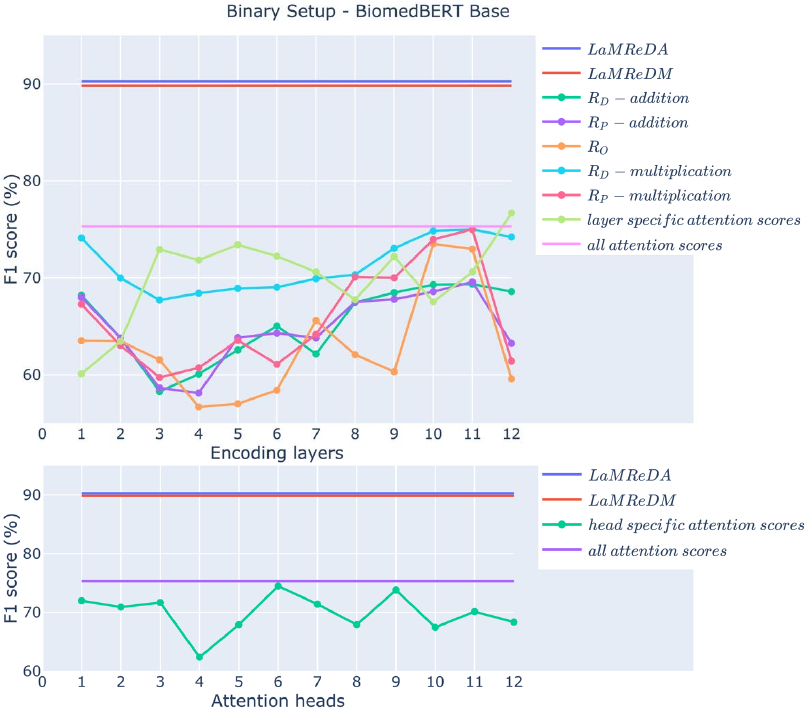}
  \caption{\label{fig:probing_redres_binary_base}ReDReS Probing (Binary setup) (BiomedBERT base): Examines LaMReDA/LaMReDM relation representations ($R_D$, $R_O$, $R_P$) and attention scores from each layer and explores average attention scores of tokens corresponding to each entity towards the other entity across attention heads. Top boundary: best LaMReDA and LaMReDM performance (Table \ref{tab:LaMReDA_LaMReDM_results}). Second boundary: classifier with average attention scores across all layers as input.}
\end{figure}

Fig. \ref{fig:probing_redres_binary_base} shows the results of probing experiments in the binary setup using ReDReS and BiomedBERT base. The 10\textsuperscript{th} and 11\textsuperscript{th} layers provide the most informative representations for relation types ($R_D$, $R_O$, and $R_P$). The $R_D$ representation, using element-wise multiplication, outperforms other representations in intra-layer comparisons, suggesting its effectiveness without end-to-end training. However, as highlighted in Section \ref{sec:results}, inter-model comparisons indicate that the transformer layers and projection layer $l()$ capture crucial information for relation detection, regardless of the aggregation function. Using context vectors with $R_O$ and $R_P$ generally offers no advantage, though $R_O$ from the 10\textsuperscript{th} and 11\textsuperscript{th} layers performs well, indicating possibly meaningful localized context. Attention scores between entities in the 12\textsuperscript{th} layer yield the best performance, surpassing the baseline of using scores from all layers, indicating strong attention between the entities in the last layer. Fig. \ref{fig:probing_redres_binary_base} reveals that the 6\textsuperscript{th} and 9\textsuperscript{th} attention heads are most informative for relation detection. Additional probing experiments are presented in Appendix \ref{sec:additional_probing_appendix}.

\section{Conclusion}
This paper presents an open-source framework for disease knowledge discovery from raw text. Facilitating further research, we contribute two new annotated datasets\footnote{All the recruited medical experts provided informed consent before participating in the annotation process. The compensation provided to the annotators was adequate, considering their demographic, particularly their country of residence.} for RS (ReDReS) and AD (ReDAD) and the respective distantly supervised datasets DiSReDReS and DiSReDAD to promote weakly supervised scenarios. Extensive evaluation explores various methods for representing relations and entities, yielding insights into optimal modeling approaches for semantic relation detection, and emphasizing language models' potential in knowledge discovery. In future work, additional existing and future LLMs can be evaluated for the task proposed in this paper.

\appendices

\section{\break Data Pipeline: Additional Information}
\label{sec:data_pipeline_appendix}
\begin{figure*}[!t]
  \centering
  \includegraphics[width=\textwidth]{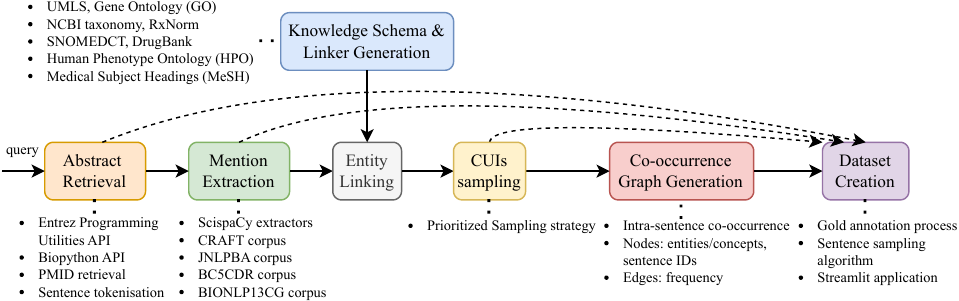}
  \caption{\label{fig:data_pipeline_scispacy}The pipeline starts with abstract retrieval using a natural language query. Next, entities are detected using the 4 different ScispaCy extractors (CRAFT, JNLPBA, BC5CDR, BIONLP13CG). The entity linking is executed, utilizing the knowledge schema and linker generation step that creates the updated linkers tailored to a range of knowledge schemes. To address the scenario of multiple concept unique identifiers (CUIs) due to the utilization of multiple linkers, we select the most relevant CUIs, using a prioritized sampling strategy. Then, the co-occurrence graph generation step models the intra-sentence co-occurrence of the extracted entities. The final step is the dataset creation using the processed text and co-occurrence graph.}
\end{figure*}

In addition to the MetaMapLite-based pipeline, we propose a second pipeline based on ScispaCy (Apache License 2.0) (Fig. \ref{fig:data_pipeline_scispacy}). The difference lies in the selection of entity extractors and linkers that map the extracted entities to the knowledge schemes. Other named entity recognition (NER) and linking tools, such as MedCAT \cite{kraljevic2021multi}, can also be incorporated. Unlike MetaMapLite, which adopts an integrated approach in which mention extraction and linking are performed simultaneously in a single step and focuses on UMLS mapping, allowing for more precise and targeted extraction of entities, ScispaCy serves a broader range of Natural Language Processing (NLP) tasks. After the retrieval of the abstracts described in Section \ref{sec:data_pipeline}, the following steps are executed: \textit{Knowledge schema and linker generation}, \textit{Mention extraction}, \textit{Entity linking} and \textit{Sampling of linked identifiers}.

\noindent
\textbf{Knowledge schema and linker generation.} ScispaCy \cite{neumann-etal-2019-scispacy} harnesses an older version of UMLS (2020AA). This version serves as the foundation upon which ScispaCy trains and constructs its linkers that operate on a char-3grams string overlap-based search mechanism, facilitating efficient and accurate entity recognition and linking processes. Following the paradigm of ScispaCy, we provide scripts for generating updated linkers tailored to a range of knowledge schemes. These include UMLS \cite{bodenreider2004unified}, Gene Ontology (GO) \cite{gene2004gene}, National Center for Biotechnology Information (NCBI) taxonomy \cite{schoch2020ncbi}, RxNorm \cite{nelson2011normalized}, SNOMED Clinical Terms (SNOMEDCT\textunderscore US) \cite{stearns2001snomed}, Human Phenotype Ontology (HPO) \cite{kohler2021human}, Medical Subject Headings (MeSH) \cite{lipscomb2000medical} DrugBank \cite{knox2024drugbank} and Gold Standard Drug Database (GS)\footnote{\url{https://www.nlm.nih.gov/research/umls/sourcereleasedocs/current/GS/index.html}}. Of particular note is the inclusion of UMLS, a unified system encompassing various knowledge bases, vocabularies, taxonomies, and ontologies pertinent to the biomedical domain. Any supported linker maps the concepts to UMLS CUIs enhancing the standardization of medical terminology. Notably, the flexibility of ScispaCy's implementation allows seamless expansion to incorporate additional knowledge bases, thereby enhancing its versatility and applicability across diverse research needs.

\noindent
\textbf{Mention extraction.} ScispaCy boasts four distinct entity extractors, each trained on different corpora, collectively encompassing a range of entity types. These extractors include NER models trained on the CRAFT corpus (with 6 entity types) \cite{bada2012concept}, JNLPBA corpus (with 5 entity types) \cite{collier-kim-2004-introduction}, BC5CDR corpus (with 2 entity types) \cite{li2016biocreative}, and BIONLP13CG corpus (with 16 entity types) \cite{kim-etal-2013-genia}. To maximize the range of the entity extraction, we leverage these diverse extractors in tandem, allowing us to capture mentions of 18 unique entity types (gene or protein, cell, chemical, organism, disease, organ, DNA, RNA, tissue, cancer, cellular component, anatomical system, multi-tissue structure, organism subdivision, developing anatomical structure, pathological formation, organism substance, and immaterial anatomical entity).

\noindent
\textbf{Entity linking.} This process enhances the semantic understanding of the extracted entities, facilitates standardization, which is a key challenge in the biomedical field \cite{bettencourt2012creating, theodoropoulos2023representation}, and promotes interoperability with external resources by associating the entities with specific concepts in the supported knowledge schemes. Each entity is subjected to a linking process where we attempt to map it to concepts within the supported knowledge bases or vocabularies. If a match is found, the entity is assigned a unique identifier, referred to as a CUI, corresponding to the specific concept in the knowledge schema. As entities may be linked to multiple knowledge sources, we merge the extracted CUIs obtained from the different linkers. This consolidation process ensures that each entity is associated with a comprehensive set of identifiers, encompassing diverse perspectives and representations across various knowledge schemes.

\begin{figure*}[!t]
  \centering
  \includegraphics[width=\textwidth]{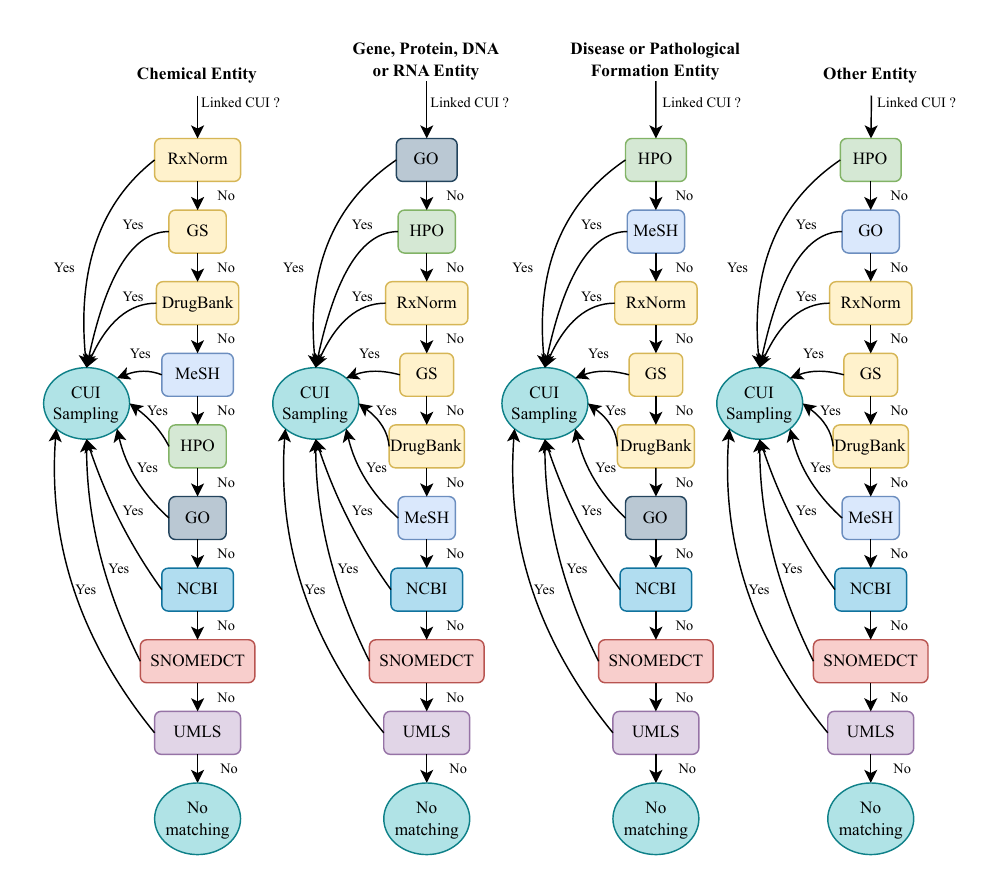}
  \caption{\label{fig:prioritized_sampling}Prioritized Concept Unique Identifier (CUI) sampling strategy: An ordered matching search for linked CUIs is designed based on the predicted type of the entity. For example, if an entity is predicted to be a chemical, the sampling strategy first checks if any linked CUIs exist in RxNorm linker, a specific knowledge schema tailored for chemicals. If linked CUIs are found, they are sampled for inclusion in the final set of CUIs associated with the entity, and the process is finished, otherwise, the search is continued in a prioritized way. The order of check is based on the potential relevance and coverage of the knowledge schema given the predicted type of the entity.}
\end{figure*}

\noindent
\textbf{Sampling of linked identifiers.} We address the scenario where multiple CUIs can be extracted for each entity due to the utilization of multiple linkers. We propose a prioritized sampling strategy (Fig. \ref{fig:prioritized_sampling}) to manage this situation and select the most relevant CUIs effectively. This strategy is designed to sample CUIs based on the predicted type of the entity (e.g., disease, gene, or chemical/drug) by prioritizing mapped CUIs from specific knowledge schemes focused on the entity being processed. For example, if an entity is predicted to be a chemical/drug, the sampling strategy first checks if any linked CUIs exist in RxNorm linker, a specific knowledge schema tailored for chemicals. If linked CUIs are found, they are sampled for inclusion in the final set of linked concepts associated with the entity, otherwise, the search is continued in a prioritized way (Fig. \ref{fig:prioritized_sampling}). We stress that the sampling strategy can be easily modified by the user based on the requirements of the research or the application.

\noindent
The \textit{co-occurrence graph generation} step is described in Section \ref{sec:data_pipeline}. 

\newpage
\noindent \textbf{Sentence Sampling Algorithm}. Given a set of sentences with defined CUIs $sent\_c$ and the co-occurrence frequency graph $co\_g$, sample $n$ number of sentences (Alg. 1). Initialize a dictionary $f\_d$ and for each sentence save the extracted CUIs pairs $c\_p$ ($extract\_conc(sent)$), the frequencies $f\_p$ of each pair extracted from the $co\_g$ ($extract\_freq(c\_p, co\_g)$) and the summation of the frequencies $t\_f$. Retrieve the sentence ids,  summed frequencies, and the inverted summed frequencies from the dictionary and append them in $ids$, $f\_l$, and $inv\_f\_l$ lists respectively. Calculate the total sums of the frequencies $t\_f\_sum$, $inv\_t\_f\_sum$ and then utilize them to define the probability distributions $\mathcal{P}$ and $\mathcal{IP}$. Sample 50\% of the sentences from $\mathcal{P}$ ($sample(\mathcal{P}, n/2)$) and 50\% from $\mathcal{IP}$ ($sample(\mathcal{IP}, n/2)$) to ensure a balance of common and potentially novel pairs of co-occurred entities in the dataset.

\begin{figure}[!t]
  \centering
  \includegraphics[width=0.85\columnwidth]{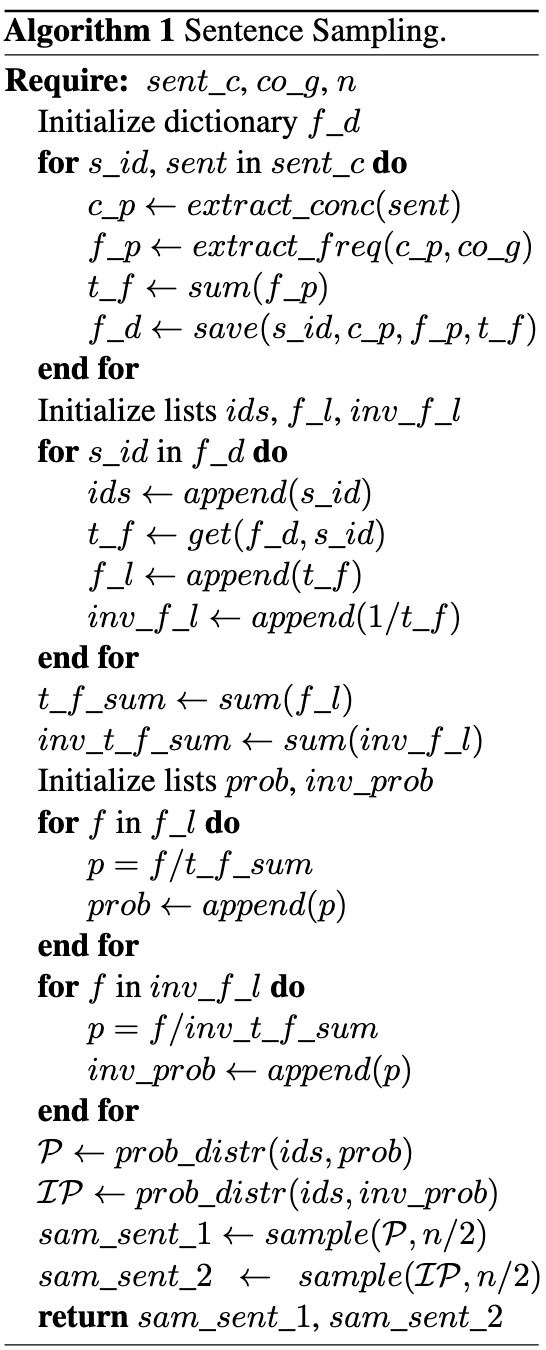}
\end{figure}

\section{\break Annotation Portal}
\begin{figure}[!h]
  \centering
  \includegraphics[width=\columnwidth]{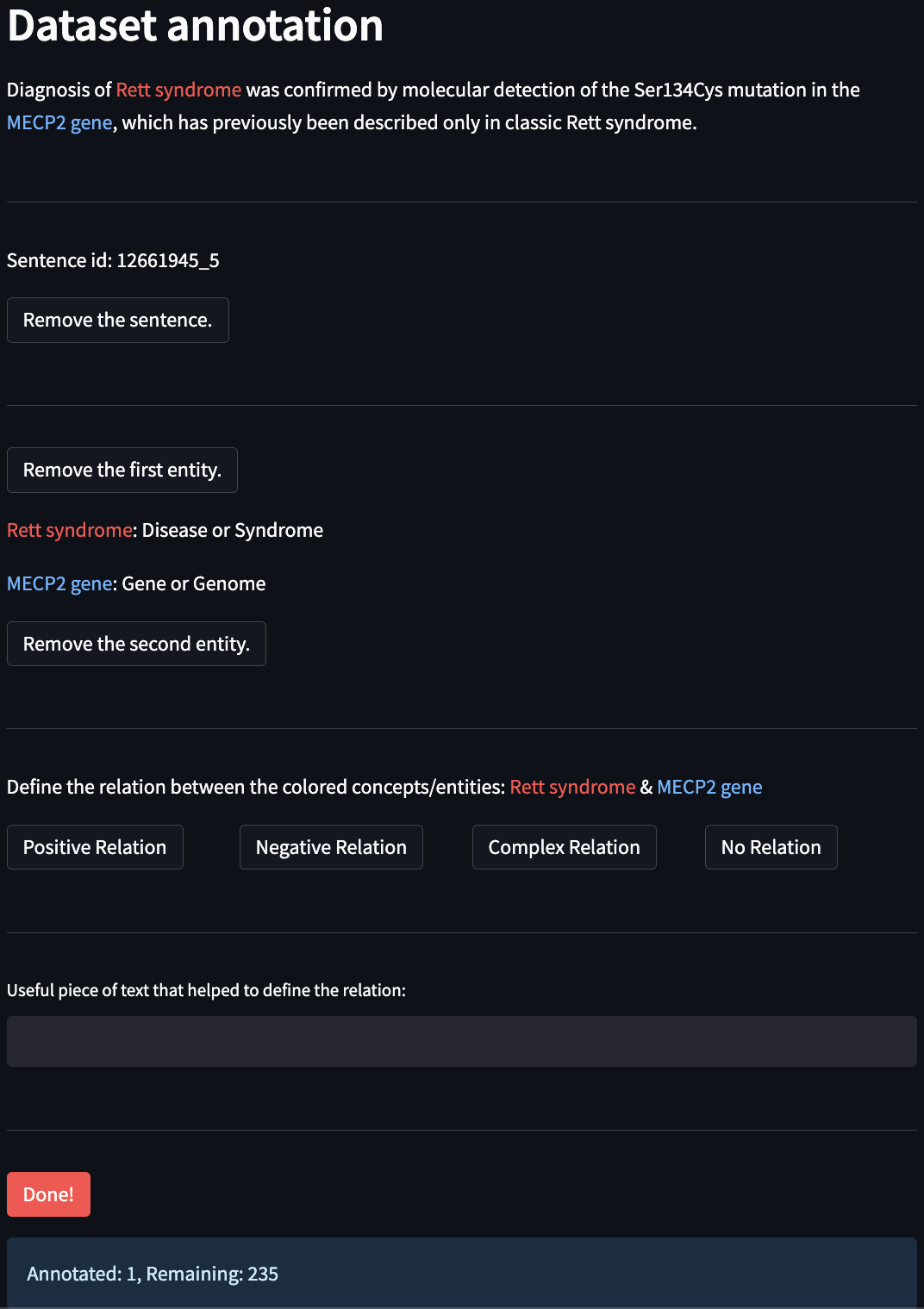}
  \caption{\label{fig:annotation_portal}Annotation portal: The annotator should define the semantic relation (\textit{Positive Relation}, \textit{Negative Relation}, \textit{Complex Relation}, and \textit{No Relation})  between the two highlighted entities: \textit{Rett syndrome} and \textit{MECP2 gene}. The annotator can remove the sentence if it is not informative and the entities if the entity detection or the type/span is incorrect. Additionally, the annotator can provide feedback, adding text that is useful for defining the relationship between the entities.}
\end{figure}

Fig. \ref{fig:annotation_portal} presents the annotation portal with an example from the ReDReS dataset. The annotator's task is to identify the semantic relation between the two highlighted entities, classifying it as either a \textit{Positive Relation}, \textit{Negative Relation}, \textit{Complex Relation}, or \textit{No Relation}. If a sentence is considered uninformative or if there are errors in entity detection, type, or span, the annotator can remove the sentence or entities. Furthermore, the annotator is encouraged to provide feedback, including any additional text that can clarify or elaborate on the relationship between the entities. By providing this supplementary information, annotators can contribute to a richer and more nuanced understanding of the relations within the data.

\newpage
\section{\break Dataset Instances}
In this section, we present some instances of different relation types in the datasets. In each example, we highlight the two detected entities.

\noindent \textbf{Positive Relation:}
\begin{itemize}
    \item \textbf{Amyloid fibrils} are found in many fatal neurodegenerative diseases such as Alzheimer's disease, Parkinson's disease, \textbf{type II diabetes}, and prion disease.
    \item AChE has become an important drug target because partial inhibition of AChE results in modest increase in ACh levels that can have therapeutic benefits, thus \textbf{AChE inhibitors} have proved useful in the \textbf{symptomatic treatment} of Alzheimer's disease.
\end{itemize}

\noindent \textbf{Complex Relation:}
\begin{itemize}
    \item When the brain's antioxidant defenses are overwhelmed by IR, it produces an abundance of reactive oxygen species (ROS) that can lead to \textbf{oxidative stress}, mitochondrial dysfunction, loss of synaptic plasticity, altered neuronal structure and microvascular impairment that have been identified as early signs of neurodegeneration in Alzheimer's disease, Parkinson's, amyotrophic lateral sclerosis, \textbf{vascular dementia} and other diseases that progressively damage the brain and central nervous system.
    \item \textbf{Autophagy inhibitor 3-methyladenine} (3-MA) attenuated the neuroprotective effect of CA, suggesting that autophagy was involved in the \textbf{neuroprotection} of CA.
\end{itemize}

\noindent \textbf{Negative Relation:}
\begin{itemize}
    \item It was not observed in \textbf{synaptopodin-deficient mice}, which lack \textbf{spine apparatus organelles}.
    \item Furthermore, the use of some kinds of \textbf{antihypertensive medication}  has been suggested to reduce the incidence of \textbf{dementia} including Alzheimer's disease.
\end{itemize}

\noindent \textbf{No Relation:}
\begin{itemize}
    \item Peripheral immune cells can cross the intact BBB, CNS neurons and glia actively regulate macrophage and lymphocyte responses, and microglia are \textbf{immunocompetent}  but differ from other macrophage/dendritic cells in their ability to direct neuroprotective \textbf{lymphocyte responses}.
    \item These techniques have thus provided morphological and functional brain alterations mapping of Alzheimer's disease: on one hand \textbf{grey matter atrophy} first concerns the medial temporal lobe before extending to the temporal neocortex and then other neocortical areas; on the other hand, \textbf{metabolic alterations}  are first located within the posterior cingulate cortex and then reach the temporoparietal area as well as the prefrontal cortex, especially in its medial part.
\end{itemize}

\section{\break Localized Context Vector}
The localized context vector is computed as follows:

\begin{itemize}
    \item Extract the attention scores of the two entities in the last encoding layer of the language model.
    \item Calculate the Hadamard product of the attention vectors.
    \item Calculate the average of the Hadamard product over the attention heads.
    \item Normalize to extract the distribution over the sequence.
    \item Extract the localized context vector by multiplying the token representations of the last encoding layer with the distribution vector.
\end{itemize}
\label{sec:context_vector_appendix}

\vspace{-2mm}
\section{\break Baseline Performance: Lower Bound}
\label{sec:baseline_performance_appendix}
To establish a baseline performance (lower bound) for comparison, in contrast to the human evaluation that serves as an upper bound, we randomly assign class labels to each instance of the test set based on the prior class distribution in the training set (Tab. \ref{tab:datasets_statistics}). This simulates a classifier with no ability to learn relations between entities. We repeat this experiment 1 million times for robustness and report the average F1-score.

In the binary setup, the baseline achieves average F1-scores of 54\% (ReDReS) and 53.16\% (ReDAD). For the multi-class setup, the average macro F1-scores range from 32.05\% (micro: 32.21\%) to 32.43\% (micro: 32.33\%) for ReDReS and ReDAD, respectively. Despite the simplicity of the baseline, the low performance highlights the challenge of the task, especially in the multi-class scenario where the model needs to distinguish between nuanced semantic relations.

\begin{figure}[!hb]
  \centering
  \vspace{-2mm}
  \includegraphics[width=\columnwidth]{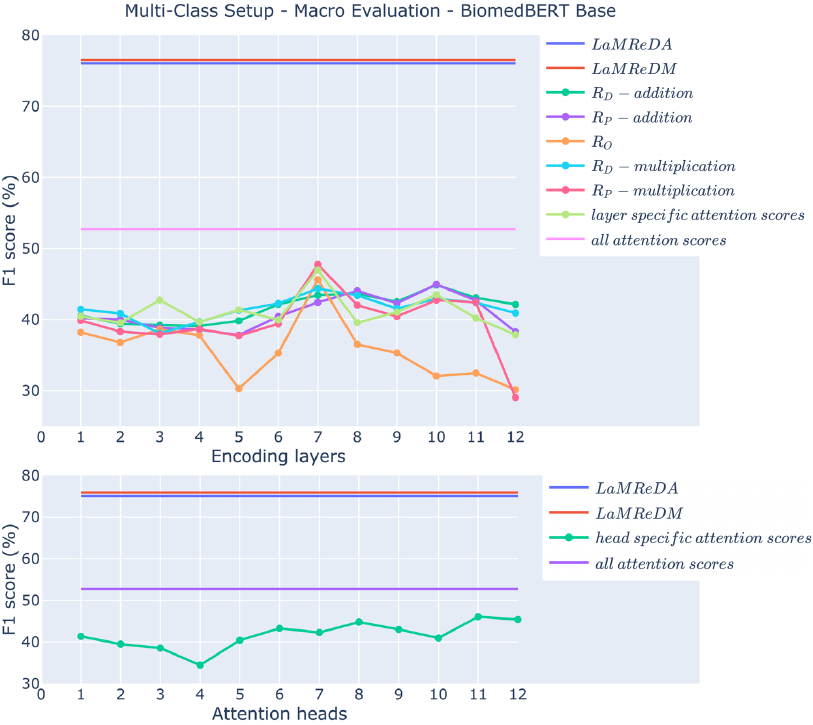}
  \vspace{-6mm}
  \caption{\label{fig:probing_redres_multi_macro_base}ReDReS Probing (Multi-class setup, Macro evaluation) (BiomedBERT base): Examines LaMReDA/LaMReDM relation representations ($R_D$, $R_O$, $R_P$) and attention scores from each layer and explores average attention scores of tokens corresponding to each entity towards the other entity across attention heads. Top boundary: best LaMReDA and LaMReDM performance (Table \ref{tab:LaMReDA_LaMReDM_results}). Second boundary: classifier with average attention scores across all layers as input.}
  \vspace{-6mm}
\end{figure}

\begin{figure}[!hb]
  \centering
  \includegraphics[width=\columnwidth]{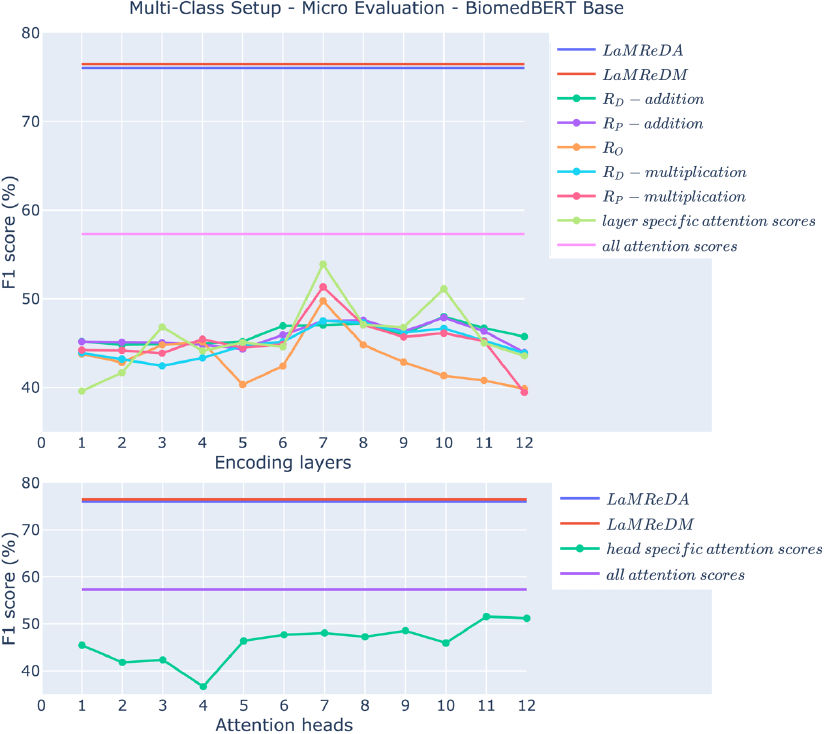}
  \vspace{-6mm}
  \caption{\label{fig:probing_redres_multi_micro_base}ReDReS Probing (Multi-class setup, Micro evaluation) (BiomedBERT base): Examines LaMReDA/LaMReDM relation representations ($R_D$, $R_O$, $R_P$) and attention scores from each layer and explores average attention scores of tokens corresponding to each entity towards the other entity across attention heads. Top boundary: best LaMReDA and LaMReDM performance (Table \ref{tab:LaMReDA_LaMReDM_results}). Second boundary: classifier with average attention scores across all layers as input.}
\end{figure}

\section{\break Probing: Additional Experiments}
\label{sec:additional_probing_appendix}

\begin{figure}[!hb]
  \centering
  \vspace{-3mm}
  \includegraphics[width=0.95\columnwidth]{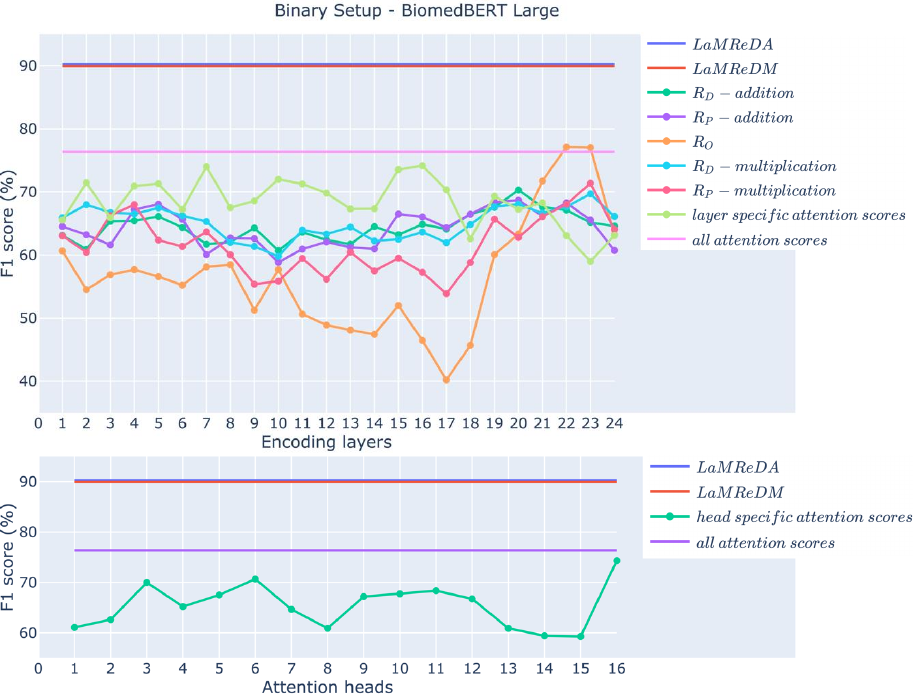}
  \vspace{-2mm}
  \caption{\label{fig:probing_redres_binary_large}ReDReS Probing (Binary setup) (BiomedBERT large): Examines LaMReDA/LaMReDM relation representations ($R_D$, $R_O$, $R_P$) and attention scores from each layer and explores average attention scores of tokens corresponding to each entity towards the other entity across attention heads. Top boundary: best LaMReDA and LaMReDM performance (Table \ref{tab:LaMReDA_LaMReDM_results}). Second boundary: classifier with average attention scores across all layers as input.}
  \vspace{-6.5mm}
\end{figure}

\begin{figure}[!hb]
  \centering
  \includegraphics[width=0.95\columnwidth]{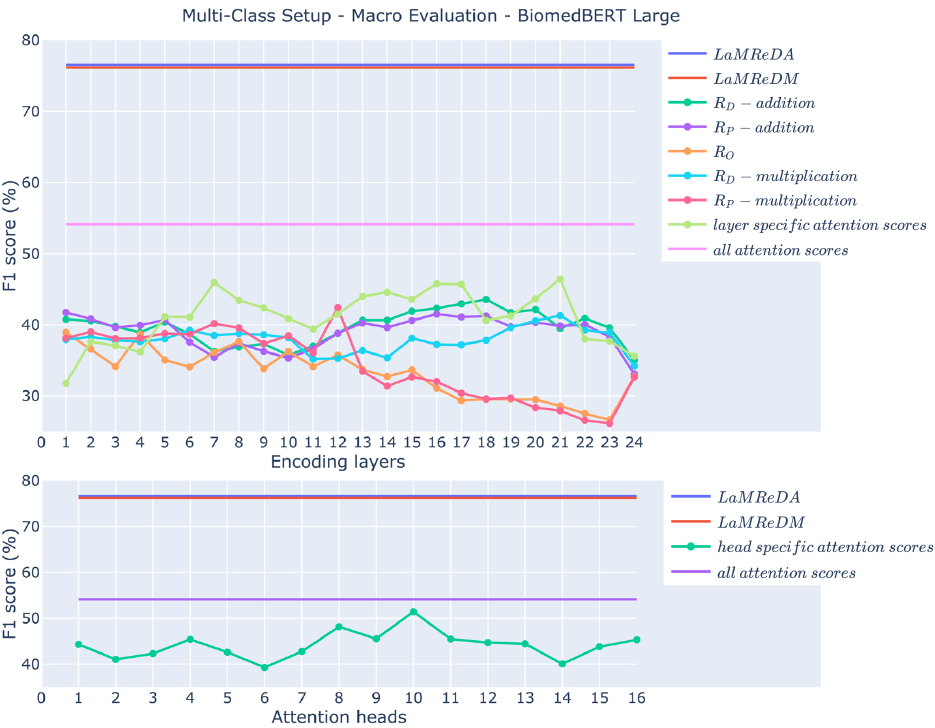}
  \vspace{-2mm}
  \caption{\label{fig:probing_redres_multi_macro_large}ReDReS Probing (Multi-class setup, Macro evaluation) (BiomedBERT large): Examines LaMReDA/LaMReDM relation representations ($R_D$, $R_O$, $R_P$) and attention scores from each layer and explores average attention scores of tokens corresponding to each entity towards the other entity across attention heads. Top boundary: best LaMReDA and LaMReDM performance (Table \ref{tab:LaMReDA_LaMReDM_results}). Second boundary: classifier with average attention scores across all layers as input.}
\end{figure}

\begin{figure}[!ht]
  \centering
  \includegraphics[width=\columnwidth]{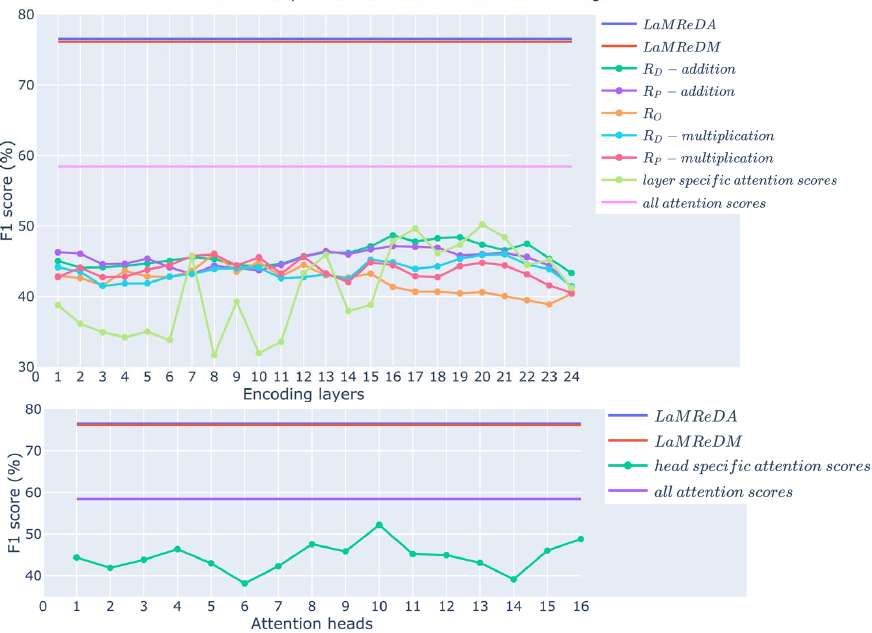}
  \caption{\label{fig:probing_redres_multi_micro_large}ReDReS Probing (Multi-class setup, Micro evaluation) (BiomedBERT large): Examines LaMReDA/LaMReDM relation representations ($R_D$, $R_O$, $R_P$) and attention scores from each layer and explores average attention scores of tokens corresponding to each entity towards the other entity across attention heads. Top boundary: best LaMReDA and LaMReDM performance (Table \ref{tab:LaMReDA_LaMReDM_results}). Second boundary: classifier with average attention scores across all layers as input.}
\end{figure}

We use the same experimental setup as described in subsection 3.3 and the experiments are conducted in the 5-fold cross-validation setting. To provide an inclusive probing analysis on ReDReS, we incorporate additional probing results in this section. Fig. \ref{fig:probing_redres_multi_macro_base} and \ref{fig:probing_redres_multi_micro_base} present the experiments in the multi-class setup using BiomedBERT base. Additionally, aiming to explore the probing capabilities of BiomedBERT large, we include the results of further experiments in Fig. \ref{fig:probing_redres_binary_large}, \ref{fig:probing_redres_multi_macro_large}, and \ref{fig:probing_redres_multi_micro_large}. These experiments investigate the model's performance in detecting semantic relations, comparing the representations and attention mechanisms at different layers and heads to understand how well the larger LM can discern complex relationships in biomedical text.

\section{\break Results with standard deviation}
\label{sec:results_with_std_appendix}

We present the results with the standard deviation when we utilize the official split of the data (Table \ref{tab:datasets_statistics}) and run the experiments 10 times with different seeds (Tables \ref{tab:lamel_diff_lms_std}, \ref{tab:LaMEL_results_std}, \ref{tab:LaMReDA_LaMReDM_results_std} and \ref{tab:lamreda_diff_lms_std}).

\begin{table}[!hb]
  \caption{\label{tab:lamel_diff_lms_std} LaMEL Results (\%) in binary setup utilizing BiomedBERT-base, BioLinkBERT-base, and BioGPT as backbone language model: Each cell shows the average F1-score from 10 runs (original test set) with the standard deviation.}
  \resizebox{\columnwidth}{!}{
  \begin{threeparttable}
        \begin{tabular}{cccc|ccc}
            \hline
            \multirow{3}{*}{\textbf{Type}\tnote{1}} & \multicolumn{3}{c}{\textbf{ReDReS}} & \multicolumn{3}{|c}{\textbf{ReDAD}}\\
            \cmidrule{2-7}
            & \textbf{BiomedBERT\tnote{2}} & \textbf{BioLinkBERT\tnote{2}} & \textbf{BioGPT} & \textbf{BiomedBERT\tnote{2}} & \textbf{BioLinkBERT\tnote{2}} & \textbf{BioGPT} \\
            \hline
            A & 90.25\textpm{0.7} & 90.68\textpm{0.9} & 89.43\textpm{1.1} & 86.73\textpm{1.7} & \textbf{87.71}\textpm{0.8} & 86.14\textpm{1.6} \\
            \hline
            D & 90.47\textpm{0.8} & 90.35\textpm{0.9} & 88.52\textpm{0.9} & 86.29\textpm{1.0} & 87.35\textpm{1.3} & 85.28\textpm{1.9} \\
            \hline
            E & \textit{90.61}\textpm{0.6} & 90.25\textpm{0.6} & \textit{89.48}\textpm{0.8} & 86.03\textpm{1.0} & 87.05\textpm{1.2} & \textit{87.42}\textpm{0.8} \\
            \hline
            H & 89.68\textpm{1.3} & \textbf{90.73}\textpm{0.6} & 89.21\textpm{0.9} & \textit{87.29}\textpm{1.0} & 87.56\textpm{1.1} & 85.55\textpm{2.0} \\
            \hline
            \hline
        \end{tabular}
    \begin{tablenotes}
      \item [1] Type of Relation Representation.
      \item [2] Base version.
    \end{tablenotes}
  \end{threeparttable}}
\end{table}

\begin{table}[!hb]
  \caption{\label{tab:LaMEL_results_std}LaMEL Results (\%) in binary setup (BiomedBERT $\square$: base, $\blacksquare$:large): Each cell shows the average F1-score from 10 runs (original test set) with the standard deviation.}
  \vspace{-1mm}
  \resizebox{\columnwidth}{!}{
  \begin{threeparttable}
        \begin{tabular}{cc|c|c|c}
            \hline
            \multirow{3}{*}{\textbf{Type}\tnote{1}} & \multicolumn{2}{c}{\textbf{ReDReS}} & \multicolumn{2}{|c}{\textbf{ReDAD}}\\
            \cmidrule{2-5}
            & \textbf{F\textsubscript{1}}\textsuperscript{$\square$} & \textbf{F\textsubscript{1}}\textsuperscript{$\blacksquare$} & \textbf{F\textsubscript{1}}\textsuperscript{$\square$} & \textbf{F\textsubscript{1}}\textsuperscript{$\blacksquare$} \\
            \hline
            A & 90.25\textpm{0.7} & 90.88\textpm{0.7} & 86.73\textpm{1.7}  & 88.90\textpm{0.7}\\
            \hline
            B & 90.29\textpm{0.5} & 90.73\textpm{0.7} & 86.89\textpm{1.1} & 88.22\textpm{1.1} \\
            \hline
            C & 90.51\textpm{0.5} & 90.71\textpm{0.6} & \textbf{87.49}\textpm{0.9} & 88.57\textpm{1.1} \\
            \hline
            D & 90.47\textpm{0.8} & \textbf{91.03}\textpm{0.6} & 86.29\textpm{1.0} & 88.22\textpm{0.9} \\
            \hline
            E & \textbf{90.61}\textpm{0.6} & 90.54\textpm{0.7} & 86.03\textpm{1.0} & 88.74\textpm{1.4}\\
            \hline
            F & 90.48\textpm{1.1} & 90.88\textpm{0.4} & 87.27\textpm{0.9} & \textbf{89.44}\textpm{0.5}\\
            \hline
            G & 90.32\textpm{0.9} & 90.35\textpm{0.7} & 86.97\textpm{1.1} & 89.12\textpm{0.5}\\
            \hline
            H & 89.68\textpm{1.3} & 90.13\textpm{1.4} & 87.29\textpm{1.0} & 88.77\textpm{1.2}\\
            \hline
            \hline
            CD{\tnote{2}} & 86.20\textpm{1.1} & 89.14\textpm{1.0} & 88.92\textpm{0.5} & 88.56\textpm{0.8}\\
            \hline
        \end{tabular}
    \begin{tablenotes}
      \item [1] Type of Relation Representation.
      \item [2] Cross-disease experiments utilizing the entity representation $E_A$: Training on ReDReS, evaluation on ReDAD, and vice versa.
    \end{tablenotes}
  \end{threeparttable}}
\end{table}

\begin{table*}[!ht]
  \caption{\label{tab:LaMReDA_LaMReDM_results_std}LaMReDA and LaMReDM Results (\%) in binary and multi-class setup (BiomedBERT $\square$: base,$\blacksquare$: large): Each cell shows the average F1-score from 10 runs (original test set) with the standard deviation.}
  \vspace{-3mm}
  \resizebox{0.96\textwidth}{!}{
  \begin{threeparttable}
        \begin{tabular}{cccc|cc|cc|cc|cc|cc}
            \hline
            \multirow{4}{*}{\textbf{Data}} & \multirow{4}{*}{\textbf{Type}\tnote{1}} & \multicolumn{4}{c}{\textbf{Binary setup}} & \multicolumn{8}{|c}{\textbf{Multi-class setup}}\\
            \cmidrule{7-14}
            & & & & & & \multicolumn{4}{c}{\textbf{Micro Evaluation}} & \multicolumn{4}{|c}{\textbf{Macro Evaluation}} \\
            \cmidrule{3-14}
            & & \multicolumn{2}{c}{\textbf{LaMReDA}} & \multicolumn{2}{|c}{\textbf{LaMReDM}} & \multicolumn{2}{|c}{\textbf{LaMReDA}} & \multicolumn{2}{|c}{\textbf{LaMReDM}} & \multicolumn{2}{|c}{\textbf{LaMReDA}} & \multicolumn{2}{|c}{\textbf{LaMReDM}} \\
            \cmidrule{3-14}
            & & \textbf{F\textsubscript{1}}\textsuperscript{$\square$} & \textbf{F\textsubscript{1}}\textsuperscript{$\blacksquare$} & \textbf{F\textsubscript{1}}\textsuperscript{$\square$} & \textbf{F\textsubscript{1}}\textsuperscript{$\blacksquare$} & \textbf{F\textsubscript{1}}\textsuperscript{$\square$} & \textbf{F\textsubscript{1}}\textsuperscript{$\blacksquare$} & \textbf{F\textsubscript{1}}\textsuperscript{$\square$} & \textbf{F\textsubscript{1}}\textsuperscript{$\blacksquare$} &
            \textbf{F\textsubscript{1}}\textsuperscript{$\square$} & \textbf{F\textsubscript{1}}\textsuperscript{$\blacksquare$} &
            \textbf{F\textsubscript{1}}\textsuperscript{$\square$} & \textbf{F\textsubscript{1}}\textsuperscript{$\blacksquare$} \\
            \hline
            \parbox[t]{2mm}{\multirow{17}{*}{\rotatebox[origin=c]{90}{ReDReS}}} & A & 90.72\textpm{0.4} & 90.74\textpm{0.7} & 90.42\textpm{1.0} & \textbf{90.71}\textpm{0.5} & 74.49\textpm{1.3} & 73.96\textpm{0.8} & 74.36\textpm{1.0} & 74.35\textpm{0.8} & 74.52\textpm{1.3} & 73.66\textpm{1.1} & 74.30\textpm{0.9} & 72.81\textpm{1.3} \\
            \cmidrule{2-14}
            & B & 90.40\textpm{0.4} & 90.28\textpm{0.9} & 90.47\textpm{0.7} & 90.06\textpm{0.6} & 74.27\textpm{1.1} & 73.72\textpm{1.3} & 74.26\textpm{1.3} & 73.57\textpm{0.9} & 74.32\textpm{1.1} & 73.65\textpm{1.1} & \textbf{74.38}\textpm{1.2} & 73.11\textpm{1.2} \\
            \cmidrule{2-14}
            & C & 90.85\textpm{0.7} & 90.75\textpm{0.8} & 90.51\textpm{0.5} & 89.14\textpm{0.8} & \textbf{74.93}\textpm{1.2} & 73.54\textpm{1.7} & 74.31\textpm{1.1} & 73.69\textpm{1.5} & \textbf{74.96}\textpm{1.1} & 73.44\textpm{1.9} & 74.10\textpm{0.9} & 73.49\textpm{0.7} \\
            \cmidrule{2-14}
            & D & 90.55\textpm{0.4} & \textbf{90.93}\textpm{0.7} & 90.61\textpm{0.4} & 90.53\textpm{0.4} & 73.61\textpm{0.6} & 73.50\textpm{1.1} & 73.02\textpm{0.9} & 73.21\textpm{0.8} & 73.71\textpm{0.8} & 73.70\textpm{1.2} & 73.24\textpm{0.5} & 73.90\textpm{1.0} \\
            \cmidrule{2-14}
            & E & 89.57\textpm{0.8} & 89.43\textpm{0.6} & 89.57\textpm{0.8} & 89.43\textpm{0.6} & 73.73\textpm{0.9} & 73.68\textpm{1.0} & 73.73\textpm{0.9} & 73.68\textpm{1.0} & 73.95\textpm{0.7} & 74.01\textpm{0.8} & 73.95\textpm{0.7} & 74.01\textpm{0.8} \\
            \cmidrule{2-14}
            & F & 90.48\textpm{0.4} & 90.62\textpm{0.8} & 90.41\textpm{0.3} & 90.43\textpm{0.8} & 72.86\textpm{1.1} & 73.82\textpm{0.9} & 73.51\textpm{0.9} & 73.33\textpm{0.9} & 72.62\textpm{0.7}& 73.94\textpm{1.2} & 73.32\textpm{1.1} & 74.50\textpm{0.8} \\
            \cmidrule{2-14}
            & G & 90.78\textpm{0.7} & 90.76\textpm{0.9} & 90.49\textpm{0.5} & 89.47\textpm{0.9} & 74.33\textpm{1.1} & 73.63\textpm{1.3} & 74.05\textpm{1.2} & 73.22\textpm{1.0} & 74.57\textpm{0.7} & 73.31\textpm{1.7} & 74.21\textpm{1.2} & 73.87\textpm{1.0} \\
            \cmidrule{2-14}
            & H & \textbf{90.91}\textpm{0.5} & 90.45\textpm{0.7} & 90.29\textpm{0.8} & 89.99\textpm{0.6} & 74.43\textpm{0.8} & 73.62\textpm{1.3} & 73.59\textpm{0.5} & 73.36\textpm{0.8} & 74.48\textpm{0.5} & 73.65\textpm{1.4} & 73.90\textpm{0.7} & 73.42\textpm{1.6} \\
            \cmidrule{2-14}
            & I & 90.86\textpm{0.4} & 90.47\textpm{0.6} & 90.62\textpm{0.6} & 89.55\textpm{0.8} & 74.75\textpm{0.9} & 73.30\textpm{0.7} & 74.29\textpm{0.7} & 74.00\textpm{1.5} & 74.80\textpm{0.8} & 73.26\textpm{1.1} & 73.88\textpm{0.8} & 73.91\textpm{1.1} \\
            \cmidrule{2-14}
            & J & 89.43\textpm{1.0} & 89.65\textpm{0.7} & 89.53\textpm{0.9} & 89.89\textpm{0.5} & 73.75\textpm{0.8} & 74.28\textpm{1.0} & 73.47\textpm{0.7} & \textbf{74.43}\textpm{1.0} & 74.05\textpm{1.1} & \textbf{75.06}\textpm{0.9} & 73.52\textpm{0.8} & \textbf{74.70}\textpm{1.3} \\
            \cmidrule{2-14}
            & K & 90.10\textpm{0.6} & 90.05\textpm{0.5} & 89.70\textpm{0.4} & 89.64\textpm{0.7} & 74.43\textpm{0.8} & 74.07\textpm{1.2} & \textbf{74.47}\textpm{0.8} & 74.38\textpm{0.6} & 74.44\textpm{0.9} & 74.23\textpm{1.5} & 74.30\textpm{0.7} & 74.02\textpm{0.9} \\
            \cmidrule{2-14}
            & L & 89.60\textpm{0.8} & 89.85\textpm{0.8} & 89.82\textpm{0.7} & 90.33\textpm{0.6} & 73.35\textpm{1.8} & \textbf{74.32}\textpm{0.7} & 73.68\textpm{1.0} & 73.90\textpm{0.9} & 73.16\textpm{1.8} & 74.04\textpm{0.6} & 73.66\textpm{1.1} & 73.52\textpm{1.2} \\
            \cmidrule{2-14}
            & M & 90.81\textpm{0.8} & 90.07\textpm{0.6} & 90.01\textpm{0.7} & 89.75\textpm{0.7} & 74.21\textpm{0.9} & 74.29\textpm{0.7} & 73.96\textpm{0.9} & 74.32\textpm{1.0} & 74.03\textpm{0.8} & 74.36\textpm{0.7} & 73.72\textpm{0.7} & 73.95\textpm{1.1} \\
            \cmidrule{2-14}
            & N & 90.73\textpm{0.7} & 90.60\textpm{0.8} & 90.72\textpm{0.8} & 90.63\textpm{0.8} & 74.55\textpm{1.2} & 73.83\textpm{1.3} & 73.86\textpm{0.9} & 73.38\textpm{1.6} & 74.66\textpm{1.3} & 73.97\textpm{1.0} & 74.13\textpm{1.0} & 73.53\textpm{0.9} \\
            \cmidrule{2-14}
            & O & 90.90\textpm{0.5} & 90.50\textpm{1.0} & \textbf{90.90}\textpm{0.5} & 90.50\textpm{1.0} & 73.99\textpm{1.1} & 73.77\textpm{1.2} & 73.99\textpm{1.1} & 73.77\textpm{1.2} & 73.83\textpm{1.2} & 73.62\textpm{1.1} & 73.83\textpm{1.2} & 73.62\textpm{1.1} \\
            \cmidrule{2-14}
            & P & 89.72\textpm{0.7} & 90.31\textpm{0.4} & 89.13\textpm{0.7} & 90.30\textpm{0.6} & 73.37\textpm{1.0} & 73.87\textpm{1.1} & 73.57\textpm{1.0} & 74.41\textpm{1.0} & 73.48\textpm{1.1} & 74.66\textpm{1.3} & 73.54\textpm{1.0} & 73.52\textpm{1.1} \\
            \cmidrule{2-14}
            \cmidrule{2-14}
            & CD{\tnote{2}} & 87.42\textpm{0.5} & 88.93\textpm{0.9} & 87.76\textpm{0.7} & 88.10\textpm{0.7} & 73.09\textpm{1.0} & 75.04\textpm{1.6} & 74.15\textpm{0.8} & 75.35\textpm{1.2} & 73.64\textpm{0.8} & 74.94\textpm{1.6} & 74.38\textpm{1.5} & 75.44\textpm{1.8} \\
            \hline
            \parbox[t]{2mm}{\multirow{17}{*}{\rotatebox[origin=c]{90}{ReDAD}}} & A & 88.31\textpm{0.8} & 89.55\textpm{0.9} & 87.98\textpm{0.8} & 89.14\textpm{0.8} & 77.64\textpm{1.4} & 79.47\textpm{1.5} & 78.34\textpm{1.3} & \textbf{80.21}\textpm{0.8} & 77.34\textpm{1.1} & 79.26\textpm{1.8} & 78.44\textpm{1.3} & \textbf{80.13}\textpm{0.8} \\
            \cmidrule{2-14}
            & B & 87.82\textpm{0.7} & 89.11\textpm{0.9} & 87.66\textpm{0.5} & 88.64\textpm{0.9} & 77.74\textpm{1.0} & 78.65\textpm{0.6} & \textbf{78.61}\textpm{0.9} & 78.91\textpm{1.1} & 77.13\textpm{1.1} & 78.58\textpm{0.6} & 77.76\textpm{0.8} & 78.98\textpm{0.8} \\
            \cmidrule{2-14}
            & C & 88.30\textpm{0.6} & 89.21\textpm{2.4} & 88.11\textpm{0.6} & 89.17\textpm{1.8} & 77.14\textpm{1.5} & 79.67\textpm{1.7} & 78.19\textpm{1.1} & 79.32\textpm{1.1} & 77.08\textpm{1.2} & 79.35\textpm{1.6} & 77.82\textpm{0.9} & 79.26\textpm{1.1} \\
            \cmidrule{2-14}
            & D & 87.33\textpm{0.9} & 89.82\textpm{0.5} & 88.18\textpm{0.8} & 88.80\textpm{2.5} & 78.28\textpm{0.9} & 79.54\textpm{1.4} & 76.81\textpm{1.2} & 78.68\textpm{1.9} & 78.26\textpm{1.0} & 78.47\textpm{1.7} & 76.67\textpm{1.2} & 78.73\textpm{1.3} \\
            \cmidrule{2-14}
            & E & 88.03\textpm{0.9} & 89.37\textpm{0.6} & 88.03\textpm{0.9} & 89.37\textpm{0.6} & 77.83\textpm{0.9} & 77.54\textpm{1.3} & 77.83\textpm{0.9} & 77.54\textpm{1.3} & 77.75\textpm{0.9} & 77.44\textpm{1.4} & 77.75\textpm{0.9} & 77.44\textpm{1.4}\\
            \cmidrule{2-14}
            & F & 87.71\textpm{0.8} & 88.45\textpm{2.9} & 87.87\textpm{0.6} & 88.54\textpm{0.6} & 77.59\textpm{1.7} & \textbf{79.74}\textpm{1.4} & 76.94\textpm{2.2} & 79.68\textpm{1.1} & 77.35\textpm{1.3} & 79.19\textpm{1.6} & 76.95\textpm{1.7} & 79.21\textpm{1.1} \\
            \cmidrule{2-14}
            & G & 88.17\textpm{0.8} & \textbf{89.83}\textpm{0.8} & 88.22\textpm{0.8} & \textbf{89.55}\textpm{0.9} & 77.83\textpm{1.5} & 79.39\textpm{1.0} & 78.13\textpm{2.2} & 79.09\textpm{1.8} & 77.5\textpm{1.3} & 79.12\textpm{1.0} & 77.88\textpm{2.4} & 78.74\textpm{1.5} \\
            \cmidrule{2-14}
            & H & 88.01\textpm{0.6} & 88.76\textpm{1.5} & 87.73\textpm{0.6} & 88.99\textpm{0.8} & 77.12\textpm{1.5} & 79.57\textpm{0.9} & 78.11\textpm{0.7} & 79.40\textpm{0.8} & 77.09\textpm{1.6} & 78.65\textpm{1.3} & 78.14\textpm{0.5} & 79.18\textpm{0.6} \\
            \cmidrule{2-14}
            & I & 87.56\textpm{2.4} & 88.05\textpm{3.9} & 87.86\textpm{1.6} & 89.45\textpm{0.8} & 77.77\textpm{1.4} & 79.23\textpm{1.1} & 78.40\textpm{1.1} &  78.99\textpm{0.9} & 77.11\textpm{1.2} & 79.05\textpm{1.1} & \textbf{78.49}\textpm{1.1} & 78.78\textpm{1.1}\\
            \cmidrule{2-14}
            & J & 87.99\textpm{0.4} & 88.89\textpm{1.5} & 87.79\textpm{0.4} & 89.06\textpm{0.9} & 77.69\textpm{0.8} & 78.40\textpm{1.1} & 77.14\textpm{3.0} & 78.40\textpm{0.9} & 77.71\textpm{0.7} & 77.94\textpm{1.2} & 76.59\textpm{2.7} & 78.18\textpm{0.8}\\
            \cmidrule{2-14}
            & K & 88.36\textpm{0.7} & 89.33\textpm{0.6} & 88.01\textpm{0.6} & 89.05\textpm{0.8} & 78.30\textpm{1.1} & 78.25\textpm{0.8} & 78.54\textpm{0.9} & 77.73\textpm{1.9} & 78.11\textpm{1.4} & 78.49\textpm{0.9} & 78.29\textpm{0.7} & 77.42\textpm{1.3}\\
            \cmidrule{2-14}
            & L & 88.25\textpm{0.6} & 89.25\textpm{0.6} & 87.87\textpm{0.6} & 89.13\textpm{0.8} & \textbf{78.48}\textpm{0.9} & 78.94\textpm{0.6} & 77.88\textpm{0.6} & 77.91\textpm{1.0} & \textbf{78.52}\textpm{0.9} & 78.16\textpm{0.9} & 77.85\textpm{0.6} & 77.37\textpm{1.4}\\
            \cmidrule{2-14}
            & M & \textbf{88.42}\textpm{0.5} & 89.57\textpm{0.5} & 88.12\textpm{0.7} & 89.40\textpm{0.6} & 77.00\textpm{1.8} & 78.85\textpm{0.8} & 78.02\textpm{1.2} & 77.12\textpm{0.7} & 76.62\textpm{2.0} & \textbf{79.66}\textpm{1.0} & 78.02\textpm{1.2} & 77.08\textpm{1.0}\\
            \cmidrule{2-14}
            & N & 87.98\textpm{0.8} & 88.94\textpm{0.9} & 88.08\textpm{0.5} & 89.47\textpm{0.5} & 78.21\textpm{0.8} & 78.92\textpm{0.9} & 78.03\textpm{1.1} & 78.78\textpm{1.6} & 78.07\textpm{0.9} & 77.91\textpm{0.7} & 77.74\textpm{1.0} & 78.60\textpm{1.4}\\
            \cmidrule{2-14}
            & O & 88.27\textpm{0.8} & 87.71\textpm{2.8} & 88.27\textpm{0.8} & 87.71\textpm{2.8} & 77.08\textpm{1.4} & 78.96\textpm{1.6} & 77.08\textpm{1.4} & 78.96\textpm{1.8} & 76.78\textpm{1.7} & 79.03\textpm{1.6} & 76.78\textpm{1.7} & 79.03\textpm{1.6}\\
            \cmidrule{2-14}
            & P & 88.02\textpm{1.0} & 88.86\textpm{1.2} & \textbf{88.33}\textpm{1.0} & 89.51\textpm{1.0} & 78.43\textpm{0.8} & 79.38\textpm{1.4} & 78.44\textpm{0.7} & 79.12\textpm{1.1} & 78.37\textpm{0.9} & 79.44\textpm{1.2} & 78.04\textpm{1.2} & 78.91\textpm{1.1}\\
            \cmidrule{2-14}
            \cmidrule{2-14}
            & CD{\tnote{2}} & 88.40\textpm{0.5} & 89.33\textpm{0.8} & 89.01\textpm{0.9} & 89.16\textpm{0.6} & 73.69\textpm{1.8} & 74.29\textpm{1.8} & 72.67\textpm{2.1} & 72.81\textpm{1.4} & 74.13\textpm{1.9} & 74.82\textpm{1.7} & 72.91\textpm{1.7} & 73.76\textpm{1.6} \\
            \hline
        \end{tabular}
    \begin{tablenotes}
      \item [1] Type of Relation Representation.
      \item [2] Cross-disease experiments utilizing the relation representation $R_A$: Training on ReDReS, evaluation on ReDAD, and vice versa.
    \end{tablenotes}
  \end{threeparttable}}
  \vspace{-2.5mm}
\end{table*}

\begin{table*}[!ht]
  \caption{\label{tab:lamreda_diff_lms_std} LaMReDA Results (\%) in binary and multi-class setup utilizing BiomedBERT-base, BioLinkBERT-base, and BioGPT as backbone language model: Each cell shows the average F1-score from 10 runs (original test set) with the standard deviation.}
  \vspace{-3.5mm}
  \resizebox{0.96\textwidth}{!}{
  \begin{threeparttable}
        \begin{tabular}{ccccc|ccc|ccc}
            \hline
            \multirow{3}{*}{\textbf{Data}} & \multirow{3}{*}{\textbf{Type}\tnote{1}} & \multicolumn{3}{c}{\textbf{Binary setup}} & \multicolumn{6}{|c}{\textbf{Multi-class setup}}\\
            \cmidrule{6-11}
            & & & & & \multicolumn{3}{c}{\textbf{Micro Evaluation}} & \multicolumn{3}{|c}{\textbf{Macro Evaluation}} \\
            \cmidrule{3-11}
            & & \textbf{BiomedBERT\tnote{2}} & \textbf{BioLinkBERT\tnote{2}} & \textbf{BioGPT} & \textbf{BiomedBERT\tnote{2}} & \textbf{BioLinkBERT\tnote{2}} & \textbf{BioGPT} & \textbf{BiomedBERT\tnote{2}} & \textbf{BioLinkBERT\tnote{2}} & \textbf{BioGPT}\\
            \hline
            \parbox[t]{2mm}{\multirow{6}{*}{\rotatebox[origin=c]{90}{ReDReS}}} & A & 90.72\textpm{0.4} & \textbf{91.15}\textpm{0.4} & \textit{90.15}\textpm{0.5} & \textbf{74.49}\textpm{1.3} & 74.06\textpm{0.9} & 72.22\textpm{1.2} & 74.52\textpm{1.3} & 74.11\textpm{1.0} & 71.80\textpm{1.2} \\
            \cmidrule{2-11}
            & D & 90.55\textpm{0.4} & 90.35\textpm{1.1} & 89.76\textpm{0.9} & 73.61\textpm{0.6} & 74.12\textpm{0.9} & 71.42\textpm{0.7} & 73.71\textpm{0.8} & 74.12\textpm{0.7} & 72.48\textpm{0.9} \\
            \cmidrule{2-11}
            & E & 89.57\textpm{0.8} & 89.98\textpm{0.6} & 89.49\textpm{0.6} & 73.73\textpm{0.9} & 73.22\textpm{0.6} & 70.17\textpm{1.1} & 73.95\textpm{0.7} & 73.38\textpm{1.1} & 69.97\textpm{0.6} \\
            \cmidrule{2-11}
            & F & 90.48\textpm{0.4} & 90.72\textpm{0.5} & 89.68\textpm{0.8} & 72.86\textpm{1.1} & \textit{74.22}\textpm{0.8} & 71.36\textpm{1.1} & 72.62\textpm{0.7} & 74.13\textpm{0.7} & 72.00\textpm{1.4} \\
            \cmidrule{2-11}
            & G & 90.78\textpm{0.7} & 90.57\textpm{0.6} & 89.93\textpm{0.7} & 74.33\textpm{1.1} & 74.00\textpm{1.5} & \textit{72.68}\textpm{1.1} & \textbf{74.57}\textpm{0.7} & \textit{74.32}\textpm{1.6} & \textit{72.48}\textpm{0.7} \\
            \cmidrule{2-11}
            & O & \textit{90.90}\textpm{0.5} & 90.75\textpm{0.9} & 89.51\textpm{0.9} & 73.99\textpm{1.1} & 74.00\textpm{1.5} & 66.48\textpm{4.9} & 73.83\textpm{1.2} & 73.69\textpm{1.7} & 67.43\textpm{2.4} \\
            \hline
            \parbox[t]{2mm}{\multirow{6}{*}{\rotatebox[origin=c]{90}{ReDAD}}} & A & \textit{88.31}\textpm{0.8} & 88.73\textpm{1.0} & 87.49\textpm{0.9} & 77.64\textpm{1.4} & 78.40\textpm{0.8} & 76.04\textpm{1.4} & 77.34\textpm{1.1} & 77.68\textpm{0.8} & 76.45\textpm{0.8} \\
            \cmidrule{2-11}
            & D & 87.33\textpm{0.9} & \textbf{88.74}\textpm{0.9} & 87.56\textpm{1.0} & \textit{78.28}\textpm{0.9} & 78.27\textpm{0.7} & \textit{76.49}\textpm{0.6} & \textbf{78.26}\textpm{1.0} & 77.89\textpm{1.4} & 76.25\textpm{1.1} \\
            \cmidrule{2-11}
            & E & 88.03\textpm{0.9} & 88.43\textpm{0.7} & \textit{87.71}\textpm{0.7} & 77.83\textpm{0.9} & \textbf{78.96}\textpm{0.9} & 75.35\textpm{1.1} & 77.75\textpm{0.9} & \textit{78.00}\textpm{0.7} & 75.36\textpm{1.1} \\
            \cmidrule{2-11}
            & F & 87.71\textpm{0.8} & 88.62\textpm{0.9} & 87.63\textpm{0.5} & 77.59\textpm{1.7} & 77.43\textpm{0.8} & 76.34\textpm{1.3} & 77.35\textpm{1.3} & 77.48\textpm{0.8} & 75.47\textpm{1.1} \\
            \cmidrule{2-11}
            & G & 88.17\textpm{0.8} & 88.72\textpm{0.8} & 87.55\textpm{0.8} & 77.83\textpm{1.5} & 77.76\textpm{1.0} & 76.35\textpm{1.1} & 77.50\textpm{1.3} & 77.33\textpm{1.1} & \textit{76.61}\textpm{0.6} \\
            \cmidrule{2-11}
            & O & 88.27\textpm{0.8} & 88.58\textpm{0.6} & 86.19\textpm{1.5} & 77.08\textpm{1.4} & 76.89\textpm{0.9} & 71.69\textpm{3.2} & 76.78\textpm{1.7} & 76.62\textpm{1.3} & 70.57\textpm{3.4} \\
            \hline
        \end{tabular}
    \begin{tablenotes}
      \item [1] Type of Relation Representation.
      \item [2] Base version.
    \end{tablenotes}
  \end{threeparttable}}
\end{table*}

\section{\break Detailed Annotation Guidelines}

\noindent \textbf{General Instructions}:
\begin{itemize}
    \item \textit{Use Sentence Information Only}: Base your annotation solely on the information provided within the sentence. Do not use external knowledge or prior information.
    \item \textit{Entity Check}: Examine the entities and their types. If an entity is incorrect, if the entity span is inaccurate (including irrelevant words), or if the entity type is incorrect (e.g., "Rett syndrome" categorized as part of the human body), click the "Remove First Entity" or "Remove Second Entity" button, corresponding to the error.
    \item \textit{Removing a Sentence}: If a sentence lacks informative content, you have the option to remove it. Use this option if you are confident that the sentence is uninformative.
\end{itemize}

\noindent \textbf{Relation Categories}:
\begin{itemize}
    \item \textit{No Relation}: Use this label if there's no semantic relation between the entities in the sentence.
    \item \textit{Positive Relation}: The two entities are directly, semantically connected.
    \item \textit{Negative Relation}: The two entities are negatively correlated. This is a rare case, and negative words or phrases (e.g., "no," "absence") often indicate this.
    \item \textit{Complex Relation}: Entities are related but not straightforwardly positive or negative. Complex reasoning might be needed to determine the semantic relation.
\end{itemize}

\noindent \textbf{Annotation Process}:
\begin{enumerate}
    \item When presented with a pair, choose the relevant relation category label.
    \item If you change your choice, you can adjust it by clicking a new button corresponding to the revised label. 
    \item Important: Once you press "Done", the instance can't be retrieved, so ensure your decision is accurate.
    \item Provide Relation Context: First, you need to finalize your choice for the relation labeling and then provide (if any) the related piece of text. If classifying a pair as related, specify the word or phrase in the sentence that influenced your decision. Use the text box provided and preferably copy-paste to avoid spelling errors. Press "Enter" after inputting the text to store it.
\end{enumerate}


\bibliographystyle{IEEEtran}
\bibliography{paper}

\begin{IEEEbiography}[{\includegraphics[width=1in,height=1.25in,clip,keepaspectratio]{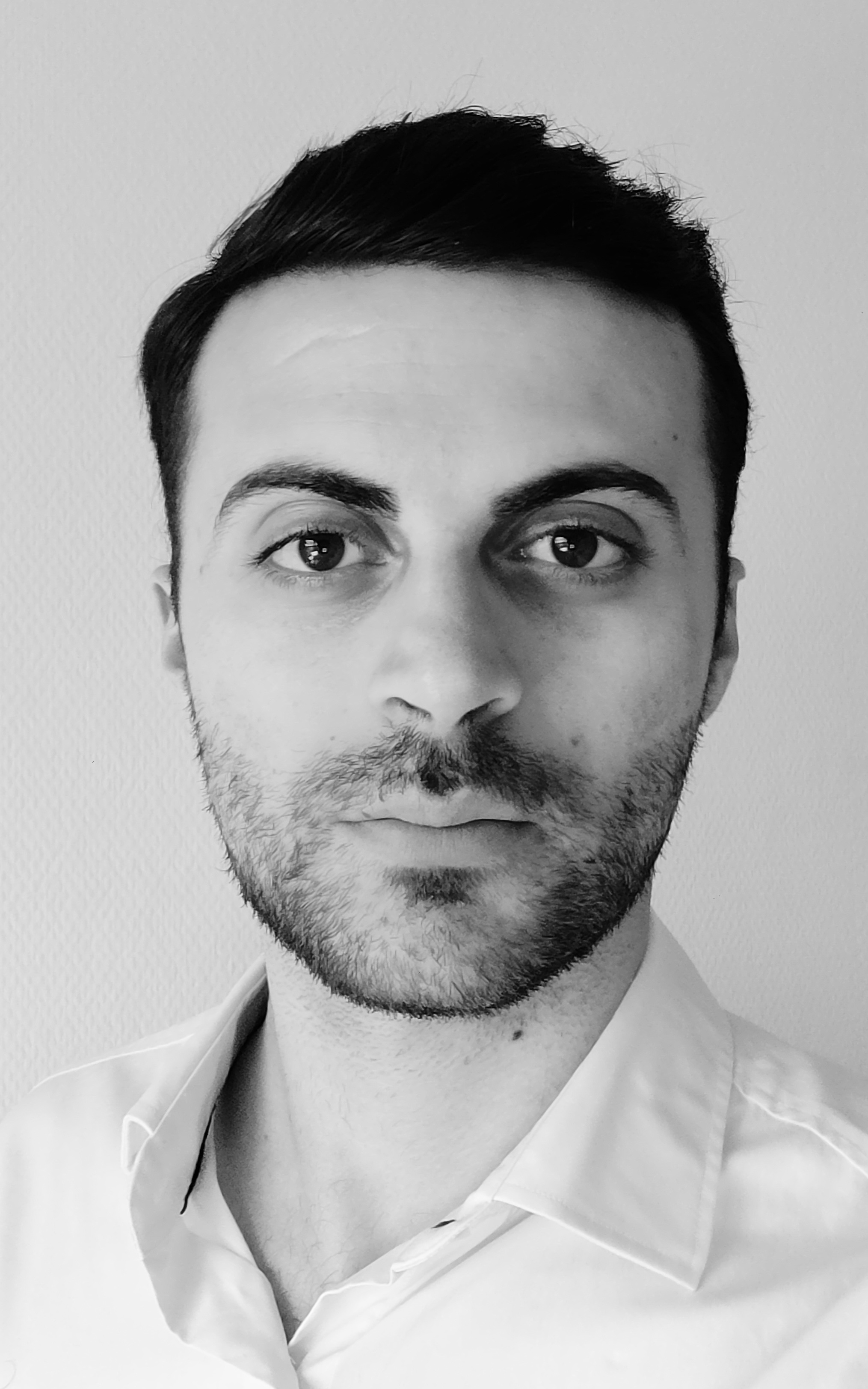}}]{Christos Theodoropoulos} received his B.Sc. and M.Sc. degrees in electrical and computer engineering from the National University of Athens and an additional advanced M.Sc. degree in computer science from KU Leuven. Currently, he is a Ph.D. Candidate in computer science at KU Leuven. His intellectual pursuits orbit at the captivating confluence of deep learning, natural language processing, knowledge discovery, and knowledge graphs, striving to invent and implement innovative approaches focused on biomedical applications. During his Ph.D., he joined IBM Research as a Research Scientist Intern, researching the person-centric knowledge graphs. Before starting his studies at KU Leuven, he worked as an Applied Researcher in the I-SENSE Group, ICCS of the National and Technical University of Athens. He serves as a Reviewer in the ACL Rolling Review (ARR) and the IEEE TRANSACTIONS ON PATTERN ANALYSIS AND MACHINE INTELLIGENCE. 
\end{IEEEbiography}

\begin{IEEEbiography}[{\includegraphics[width=1in,height=1.25in,clip,keepaspectratio]{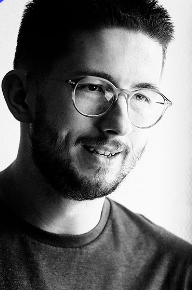}}]{Andrei Catalin Coman} received his B.Sc. and M.Sc. degrees from the University of Trento. He is a Research Assistant at the Idiap Research Institute and a Ph.D. Candidate in electrical engineering at the École Polytechnique Fédérale de Lausanne (EPFL). His research leverages deep learning techniques at the intersection of natural language processing and knowledge graphs, aiming to develop innovative techniques for knowledge representation and language understanding. Previously he worked as an Applied Scientist Intern in Amazon Science, being part of the Alexa AI team. Before moving to Idiap, he was a Research Assistant in the SiS Lab of the University of Trento and a Visiting Researcher in the AHC Lab of the Nara Institute of Science and Technology. Earlier he was an R\&D Technologist in the eHealth Lab of the Bruno Kessler Foundation. He serves as a Reviewer in the ACL Rolling Review (ARR).
\end{IEEEbiography}

\begin{IEEEbiography}[{\includegraphics[width=1in,height=1.25in,clip,keepaspectratio]{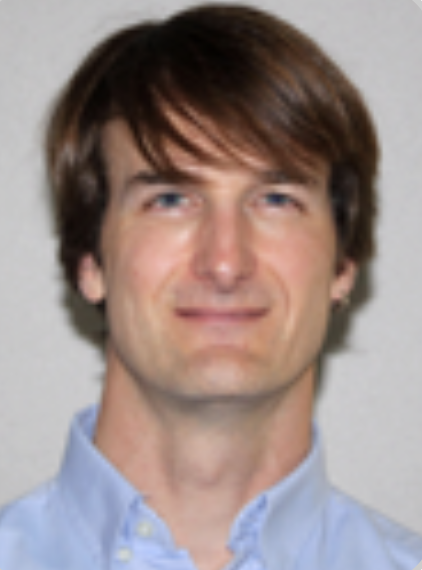}}]{James HENDERSON} received his B.Sc. degree from the Massachusetts Institute of Technology and his M.Sc. and Ph.D degrees from the University of Pennsylvania, in computer science. He is a Senior Researcher at the Idiap Research Institute, where he heads the Natural Language Understanding group. He is currently an Action Editor for the journal Transactions of the Association for Computational Linguistics (TACL), and was previously on the editorial board of the \textit{Computational Linguistics} journal.  He was Program Co-Chair for EMNLP-CoNLL 2012. His research interests are at the intersection of machine learning and natural language processing, including a long history of research in deep learning and structured prediction.
\end{IEEEbiography}

\begin{IEEEbiography}[{\includegraphics[width=1in,height=1.25in,clip,keepaspectratio]{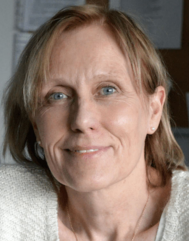}}]{MARIE-FRANCINE (SIEN) MOENS} received the M.Sc. and the Ph.D. degrees in computer science from KU Leuven. She is currently a Full Professor at KU Leuven. Her research interests include machine learning for natural language processing and the joint processing of language and visual data, deep learning, multimodal and multilingual processing, and machine learning models for structured prediction and generation. She is a fellow of the European Laboratory for Learning and Intelligent Systems (ELLIS). In 2021, she was the General Chair of the 2021 Conference on Empirical Methods in Natural Language Processing (EMNLP 2021). In 2011 and 2012, she was appointed as the Chair of the European Chapter of the Association for Computational Linguistics (EACL) and was a member of the Executive Board of the Association for Computational Linguistics (ACL). She is currently an Associate Editor of IEEE TRANSACTIONS ON PATTERN ANALYSIS AND MACHINE INTELLIGENCE and was a member of the Editorial Board of the journal Foundations and Trends in Information Retrieval, from 2014 to 2018. She is the Holder of the ERC Advanced Grant CALCULUS (2018–2024) granted by the European Research Council. From 2014 to 2018, she was the Scientific Manager of the EU COST Action iV\&L Net (The European Network on Integrating Vision and Language).
\end{IEEEbiography}

\EOD

\end{document}